\documentclass[preprint,12pt]{elsarticle}

\usepackage{amssymb}

\usepackage{enumitem} 
\usepackage[utf8]{inputenc}
\usepackage[english]{babel}
\usepackage{fancyhdr}
\usepackage{lastpage}
\usepackage{graphicx}

\usepackage{scalerel}
\usepackage{tikz}
\usetikzlibrary{shapes,arrows}
\usepackage{epstopdf}
\usetikzlibrary{shapes,arrows,positioning,fit}
\usepackage{caption}
\usepackage{subfigure}
\usepackage{tabularx}
\usepackage{amsmath}
\usepackage{mathtools}
\usepackage{multirow}
\usepackage{algorithm}
\usepackage{algpseudocode}

\journal{}

\begin{document}

	\setcounter{secnumdepth}{5}

\begin{frontmatter}


\title{Two-stage quality adaptive fingerprint image enhancement using Fuzzy c-means clustering based fingerprint quality analysis}

\author[]{Ram Prakash Sharma\corref{mycorrespondingauthor}}
\cortext[mycorrespondingauthor]{Corresponding author}
\ead{phd1501201003@iiti.ac.in}
\author[]{Somnath Dey}
\ead{somnathd@iiti.ac.in}
\address{Discipline of Computer Science and Engineering \\ Indian Institute of Technology
	Indore, India}

\begin{abstract}
	Fingerprint recognition techniques are immensely dependent on quality of the fingerprint images. To improve the performance of recognition algorithm for poor quality images an efficient enhancement algorithm should be designed. Performance improvement of recognition algorithm will be more if enhancement process is adaptive to the fingerprint quality (wet, dry or normal). In this paper, a quality adaptive fingerprint enhancement algorithm is proposed. The proposed fingerprint quality assessment algorithm clusters the fingerprint images in appropriate quality class of dry, wet, normal dry, normal wet and good quality using fuzzy c-means technique. It 
	considers seven features namely, mean, moisture, variance, uniformity, contrast, ridge valley area uniformity and ridge valley uniformity into account for clustering the fingerprint images in appropriate quality class. Fingerprint images of each quality class undergo through a two-stage fingerprint quality enhancement process. A quality adaptive preprocessing method is used as front-end before enhancing the fingerprint images with Gabor, short term Fourier transform and oriented diffusion filtering based enhancement techniques. Experimental results show improvement in the verification results for FVC2004 datasets.
	Significant improvement in equal error rate is observed while using quality adaptive preprocessing based approaches in comparison to the current state-of-the-art enhancement techniques. 
\end{abstract}


\end{frontmatter}



\section{Introduction}

In the recent era, Biometric based authentication has drawn the researcher's attention due to its uniqueness among individuals. It has been extensively deployed by many countries for the identification of persons such as US Visitor and Immigration Status Indicator Technology (US-VISIT) \cite{USVISIT2004}, India's Unique identification card by Unique Identification Authority of India (UIDAI) \cite{UIDAI}. Different biometric traits such as fingerprints, face, iris, voice, signature are used for automatic identification of an individual based on their physiological and behavioral characteristics. Fingerprint based recognition \cite{KASBAN2016,YANG2008} is one of the most reliable way of biometric authentication because of their universality, distinctiveness and accuracy. However, the quality of the fingerprint image heavily affects the performance of the recognition algorithms \cite{journals/pami/RathaKCJ96}. Therefore, it is essential to capture good quality (clear ridge and valleys) fingerprint images to improve the performance of the recognition algorithms. Due to some environmental and physical conditions of users skin, it is quite difficult to acquire a good quality fingerprint image every time. Most of the fingerprint recognition algorithm utilize minutiae points for matching \cite{maltoni2009handbook}. A poor quality image may output spurious minutiae points and ignore some genuine minutiae points which may cause degradation in recognition performance. Therefore, a robust fingerprint enhancement algorithm is required for the low-quality images resulting into accurate detection of minutiae points.

\begin{figure}[h]
	\centering
	\resizebox{0.8\textwidth}{!}{
		\includegraphics[]{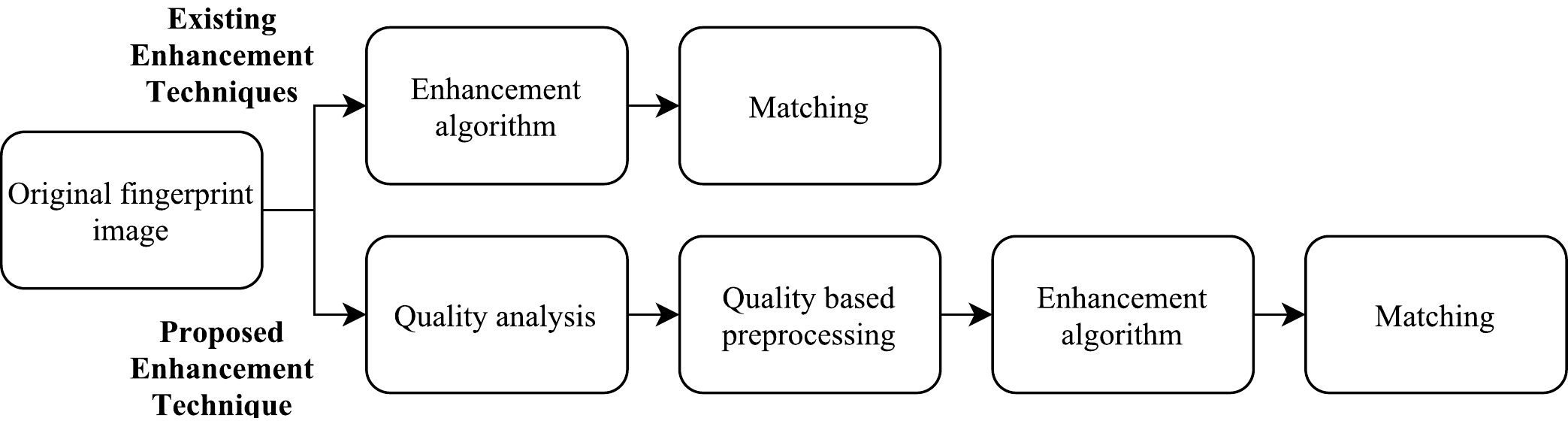}}

	\caption{Thematic diagram of the proposed approach with respect to the traditional enhancement methods}
	
	\label{thematic}
\end{figure}

There are various on-going and past efforts in the enhancement of fingerprint images \cite{OGORMAN198929,turroni2012}. Some of these techniques are based on spatial domain methods such as contextual filters \cite{OGORMAN198929}, Gabor filters and its variations \cite{journals/pami/HongWJ98,gottschlich2012curved} and oriented diffusion filtering \cite{Gottschlich2012} etc. Other enhancement methods utilizes frequency domain methods such as log-gabor filters \cite{Wang2008301}, short term fourier transform (STFT)\cite{Chikkerur2007198}, directional fourier transform and simple fourier analysis \cite{Zhu2004} etc.
A single-stage processing either in frequency or spatial domain for enhancement of low-quality fingerprint images is not enough due to the difference in ridge structure of each individual. To address this limitation some of the recent methods \cite{Yang2013,Ghafoor2014} combine both the frequency and spatial domain methods to produce better enhancement results. However, all of these methods depend on the accurate extraction of contextual information (ridge frequency and orientation). It is quite difficult to  precisely extract ridge orientation and frequency information from low-quality fingerprint images. Due to the erroneous computation of these information enhanced fingerprint image can produce some spurious minutiae points and ignores some genuine minutiae points.
Thus, a method providing adequate enhancement for high or good quality fingerprint images may not be able to give adequate enhancement for low or bad quality fingerprint images. Therefore, it is desirable that enhancement process should be adaptive to fingerprint quality. Hence, the main motive of our work is to design a quality adaptive enhancement method for fingerprint images of different characteristics. Figure \ref{thematic} shows the thematic diagram of our proposed method with respect to traditional enhancement methods. It shows that our proposed method preprocesses the fingerprint image based on its quality analysis before final enhancement. On the contrary, existing enhancement works uses similar enhancement for fingerprint images irrespective of there quality nature (good or bad, dry or wet).

In order to alleviate the limitations in the existing enhancement algorithms for low-quality (dry, wet, poor ridge valley contrast etc.) fingerprint images, a quality adaptive fingerprint enhancement scheme is proposed in this work. The proposed technique enhances the fingerprint images based on their quality class ($Q_c$) which can be dry, wet, normal dry, normal wet or good. Proposed method works in two phase. In the first phase, a fingerprint quality assessment (FQA) method is designed to cluster fingerprint images into different quality classes. In FQA, a set of features namely moisture (MI), mean (M), variance (V), ridge valley area uniformity (RVAU), ridge line count (RLC), uniformity, contrast, radial power spectrum (RPS), ridge valley uniformity (RVU), Gabor and Gabor-Shen are extracted from the fingerprint images of different quality class $Q_c$. The best feature subset among all the extracted features is obtained using a novel one-way analysis of variance (ANOVA) test. Best feature subset incorporates uniformity, contrast, mean, variance, moisture, RVAU and RVU as the most discriminative features for quality clustering of fingerprint images. The selected feature subset is fed as input to the fuzzy c-means clustering algorithm which clusters fingerprint images in different quality class. In second phase, two-stage fingerprint quality enhancement (FQE) is performed which comprises of a quality adaptive preprocessing (QAP) method followed by either Gabor, STFT or orientation diffusion filtering (ODF) enhancement technique. After FQA, each quality cluster of fingerprint images is fed to the QAP comprising of parameter adaptive unsharp masking filter, contrast limited adaptive histogram equalization (CLAHE) and gaussian smoothing. Further, other state-of-the art enhancement methods (Gabor/STFT/ODF) are utilized to enhance the fingerprint images.

The proposed fingerprint enhancement scheme is based on the quality nature of the input fingerprint images. Therefore, it is able to handle the various fingerprint quality contexts such as dry, normal dry, good, normal wet and wet. In a nutshell, the major contributions are summarized as follows:
	\vspace{-2mm}
\begin{itemize}
	\item The proposed FQA method incorporates a new set of features namely, moisture, mean, variance, RVAU and RLC with the other existing features uniformity and contrast from NFIQ 2.0 \cite{nfiq2.0} and RPS, RVU, Gabor and Gabor-Shen from Olsen et al. \cite{olsen2016}. These eleven features are used to select most influential feature set which affect the fingerprint image quality. 
	\item A novel one-way ANOVA based feature selection method to determine discriminative features which can differentiate fingerprint images of each quality class is proposed in this paper.
	\item We have proposed an objective methodology to cluster fingerprint images into different quality class using fuzzy c-means.
	\item The proposed FQE approach achieves significant performance improvement by using QAP as front-end with the Gabor/STFT/ODF fingerprint enhancement algorithms.
\end{itemize}

The rest of the paper is organized as follows: Related works on fingerprint quality enhancement and assessment are summarized in Section \ref{RelatedWork}. Section \ref{Proposed} describes the proposed method. Experimental results of FQA and fingerprint verification after QAP based enhancement are presented in Section \ref{Experiments}. Conclusions are drawn in Section \ref{Conclusions} with a glimpse of future research direction.
	\begin{figure}[h]
	\centering
	  \resizebox{1\textwidth}{!}{
	\includegraphics[]{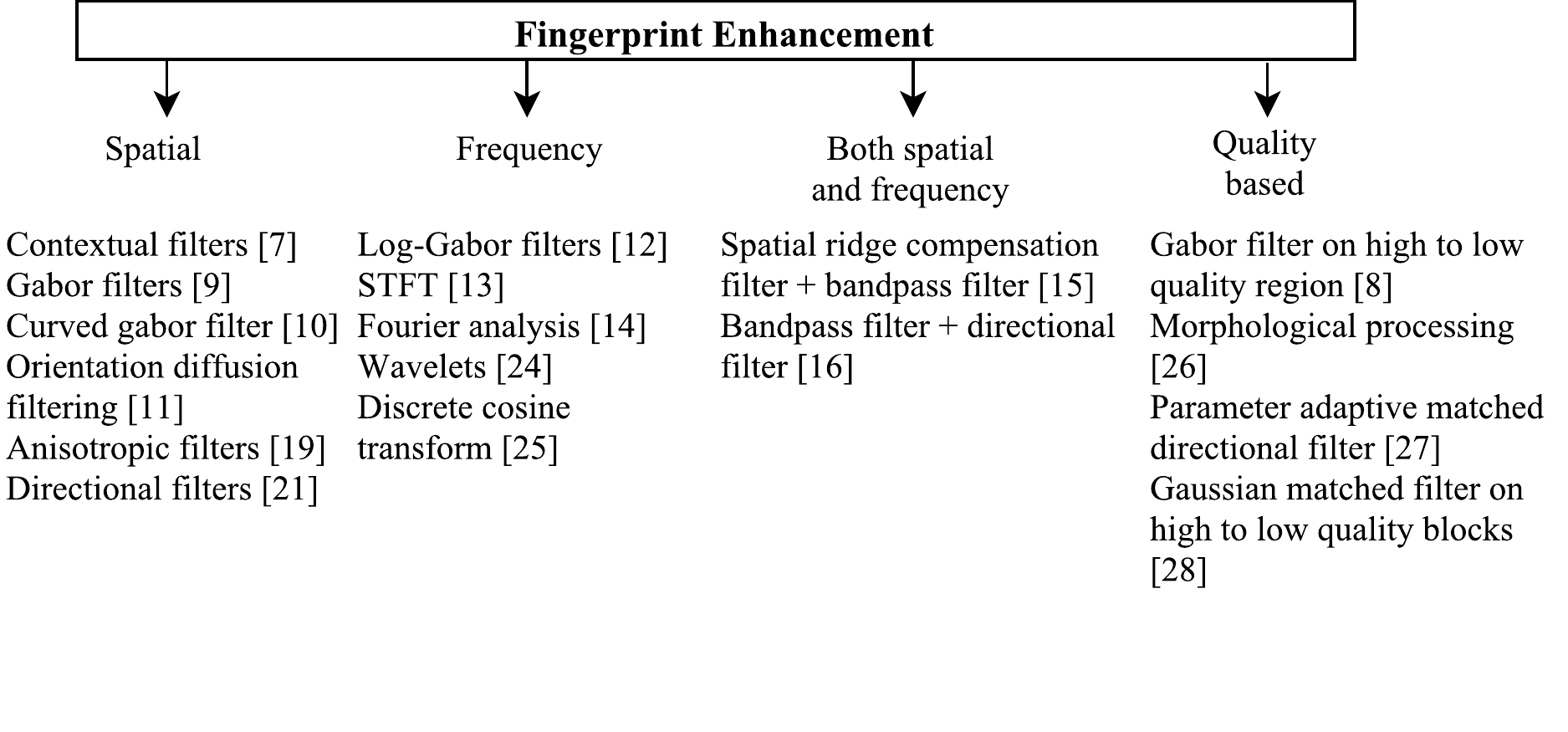}}
\vspace{-17mm}
	\caption{Features of few existing fingerprint enhancement techniques}
	
	\label{relatedwork}
\end{figure} 	

	\section{Related work}\label{RelatedWork}
	
	The existing fingerprint enhancement algorithms are roughly classified into spatial domain enhancement techniques and frequency domain enhancement techniques. Further, there are methods in the literature which utilizes both spatial and frequency domain filters for effective and precise extraction of contextual information (ridge orientation and frequency) to enhance low quality fingerprint images. Some of the fingerprint enhancement methods use quality (dry, wet or good) information of fingerprint images in combination with spatial or frequency domain filters. Therefore, all the existing enhancement methods are divided into four categories namely, spatial domain, frequency domain, both spatial and frequency domain and quality adaptive (spatial or frequency) fingerprint enhancement techniques. Salient features of the few existing fingerprint enhancement techniques is shown in Figure \ref{relatedwork}. Here, we discuss few relevant fingerprint  enhancement techniques of these categories along with some fingerprint quality analysis methods present in the literature.
	
	\subsection{Spatial domain enhancement methods}
	
	O'Gorman et. al. \cite{OGORMAN198929} proposed a technique to design contextual filters for fingerprint image enhancement. The contribution of their work is to quantify and justify the functional relationships between fingerprint image features (ridge orientation, ridge width, valley width) and filter parameters (filter mask size and filter parameters). For efficient enhancement, the filter is precomputed in 16 different directions to enhance the fingerprint image by application of oriented matched filter masks. The most popular and widely used spatial domain fingerprint image enhancement algorithm is Gabor based enhancement algorithm proposed by Hong et. al \cite{journals/pami/HongWJ98}. Gabor filter employs the frequency and orientation information for enhancement of images. Filter parameters are adapted for each block on the basis of corresponding ridge frequency and orientation to preserve the true ridge valley structure. Any error in computation of ridge orientation and ridge frequency for the blocks of fingerprint image may cause an adverse effect in the enhancement process which will influence the recognition performance. Another variation in traditional Gabor filter for fingerprint enhancement is proposed by  Gottschlich et. al. in \cite{gottschlich2012curved}, which uses curved Gabor filter for enhancing curved ridge valley structures. Curved Gabor filter locally adapts their shape to the direction of flow and enables the selection of parameters. In \cite{Gottschlich2012}, Gottschlich et. al. proposed a orientation diffusion filtering based enhancement technique. Orientation diffusion filtering utilizes local orientation of the fingerprint ridge and valley flow, followed by a locally adaptive contrast enhancement. Some other spatial domain fingerprint enhancement techniques can be found in \cite{GREENBERG2002227,YiHu2010,Fronthale2008}.

	\subsection{Frequency domain enhancement methods}
	
	Wang et al. \cite{Wang2008301} proposed a frequency domain implementation of Log-Gabor filter. Log-Gabor function effectively improves the contrast between ridges and valleys while preserving the ridge structure of fingerprint images since it does not involve any DC component. 
	Most widely used fingerprint enhancement approach in the frequency domain is proposed by Chikkerur et. al in \cite{Chikkerur2007198}. They exploit STFT analysis and contextual filtering in Fourier domain for fingerprint quality enhancement. This algorithm has an advantage that it computes all the intrinsic images (ridge orientation, ridge frequency and foreground region mask) simultaneously using STFT analysis and it utilizes full contextual information for enhancement. A top down iterative filtering based fingerprint enhancement is proposed by Zhu et. al in \cite{Zhu2004}. A new filter is formed by combining filters of \cite{Sherlock1994,Willis2001} to filter the original fingerprint image. This method initially filters the whole image and then divides it into sub-images for further enhancements using the same method until the sub image size reaches to a selected threshold. This scheme is not well suited for high curvature fingerprint images which contain fragmentary ridges.
	Other variations of the frequency domain fingerprint enhancement methods can be found in \cite{Ching2003,Jirachaweng2007}.

	\subsection{Both spatial and frequency domain enhancement methods}
	In \cite{Yang2013}, Yang et. al. proposed an effective two stage, spatial and frequency domain enhancement scheme by learning from the fingerprint images. In the first stage, a spatial ridge compensation filter is used to enhance ridge areas and contrast of local ridges. The second stage comprises a frequency bandpass filter which is separable in angular and radial frequency domain and its parameters are learned from the input images and first stage enhanced images.
	 In \cite{Ghafoor2014}, Ghafoor et. al proposed a method based on 2-fold local adaptive contextual filtering. Initially, frequency domain filtering is performed using bandpass filters which is followed by local directional filtering in the spatial domain to enhance the fingerprint image.

	\subsection{Quality adaptive enhancement methods}
	There are some fingerprint enhancement methods \cite{turroni2012,Yun2006101,bartunek2013adaptive} presented in the literature which considers quality adaptive aspects for enhancing fingerprint images.
	In \cite{turroni2012}, Turroni  et. al. proposed a method which rely on the quality of the input fingerprint image. They proposed a scheme which determines the high and low quality regions using the combined image after convolution with Gabor filter bank and local ridge flow homogeneity image. Fingerprint enhancement is done selectively using contextual filtering starting from high-quality regions to low-quality regions.
	In \cite{Yun2006101}, Yun et. al proposed a method that follows an adaptive preprocessing approach to enhance the fingerprint quality appropriately. Different morphological preprocessing stages are designed to enhance oily and dry fingerprint images. Oily or dry nature of fingerprint image is determined using five features namely, mean, variance, block directional difference, ridge-valley thickness ratio and orientation change. 
	In \cite{bartunek2013adaptive}, Bartunek et. al proposed an adaptive fingerprint enhancement method. There method is
    based on spatial contextual filtering using parameter adaptive matched directional filters. 
    In \cite{Sutthiwi2010}, Sutthiwi et. al. proposed quality diffusion based iterative fingerprint enhancement method by using Gaussian-matched filter. 
    Initially, high-quality blocks determined using signal to noise ratio are enhanced which are fed to unreliable low-quality regions for effective enhancement. However, the computational complexity of these algorithms is high because of their iterative nature.
	
	\subsection{Fingerprint quality analysis methods}\label{qualityanalysis}
	To assess dry, wet, good and normal nature of fingerprint images some fingerprint quality analysis methods \cite{lim2004,MUNIR2012,tabassi05} are present in literature. Lim et al. \cite{lim2004} proposed a block quality classification method to classify a fingerprint image blocks in very good, good, bad and very bad quality clusters. There feature vector comprises of directional strength, sinusoidal local ridge/valley pattern, ridge/valley uniformity and core occurrences of a fingerprint image sub-block.
	Tabassi et. al \cite{tabassi05} proposed a classifier based method which defines the quality as degree of separation between match and non-match distributions of a given fingerprint. It classifies the quality of fingerprint image into five levels: poor, fair, good, very good, and excellent using features vector consisting clarity of ridges and valleys, size of the image, and measure of number and
	quality of minutiae. In \cite{MUNIR2012}, Munir et. al. proposed a hierarchical k-means clustering based fingerprint quality classification method which classifies fingerprint images as good, dry, wet or normal. A set of frequency (energy concentration) and statistical features (mean, uniformity, smoothness, image inhomogeneity etc) are utilized to classify fingerprint image in different quality class. Some other fingerprint quality analysis methods can be found in \cite{Ram2017,alonso2007comparative,Bharadwaj2014}.
	
	\begin{figure}[h]
		\centering
		
		\includegraphics[width=14cm,height=10cm,keepaspectratio]{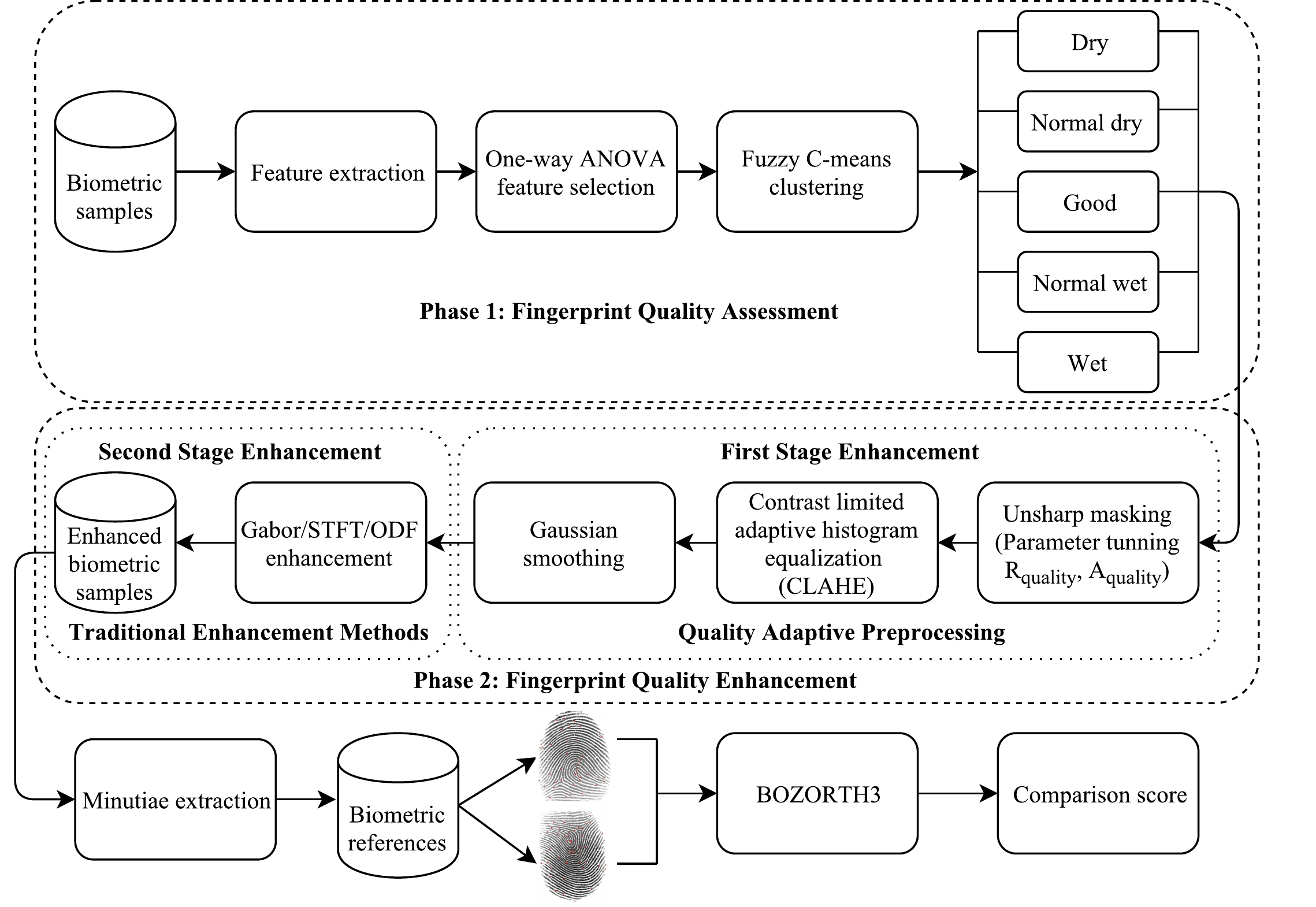}
		\caption{Block diagram of the proposed two-stage fingerprint enhancement method using fingerprint quality analysis}
		
		\label{blockdiagram}
	\end{figure} 	
	
	\vspace{-4mm}
\section{Proposed method}\label{Proposed}
Analysis of the traditional prior works \cite{journals/pami/HongWJ98,Gottschlich2012,Chikkerur2007198}, show that the existing fingerprint enhancement algorithms are not adequate in enhancing different low-quality (dry or wet etc) fingerprint images. To overcome the demerits of these methods, a new and effective quality adaptive enhancement scheme is proposed in this paper. The block diagram of the proposed method is illustrated in Figure \ref{blockdiagram}. Proposed enhancement scheme works in two phases: 1) Fingerprint quality assessment and 2) Fingerprint quality enhancement. In the first phase, a FQA algorithm is designed to cluster a fingerprint image into suitable quality class $Q_c$. The second phase comprises of two-stage FQE scheme in which the fingerprint image is pre-enhanced with quality adaptive preprocessing in the first stage. Later in the second stage, QAP based enhanced fingerprint image is fed to the Gabor/STFT/ODF enhancement techniques for further enhancement. Detailed description of each of the processing block is provided in following section.
\subsection {Fingerprint quality assessment}

\begin{figure*}[h]
	\centering
	
	\subfigure[]
	{
		\includegraphics[width=0.9in]{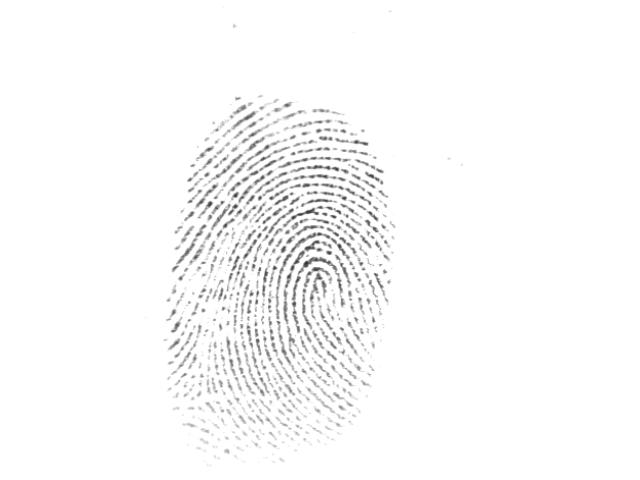}
		\label{dry}
	}
	\subfigure[]
	{
		\includegraphics[width=0.9in]{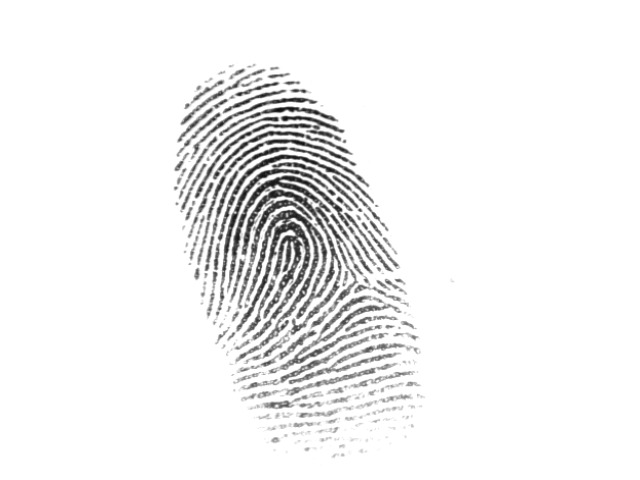}
		\label{nd}
	}		
	\subfigure[]
	{
		\includegraphics[width=0.9in]{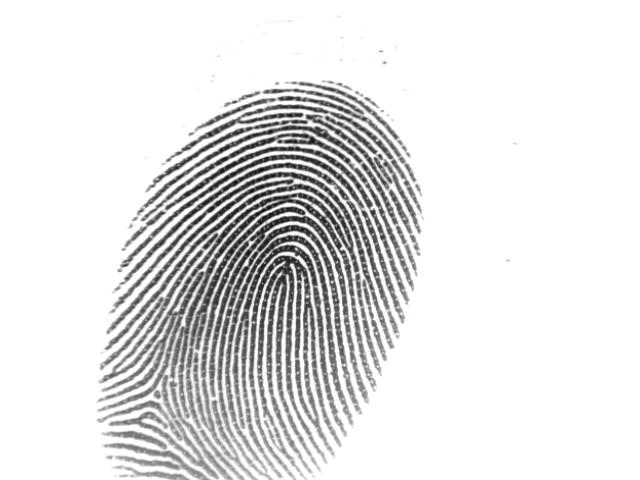}
		\label{good}
	}
	\subfigure[]
	{
		\includegraphics[width=0.9in]{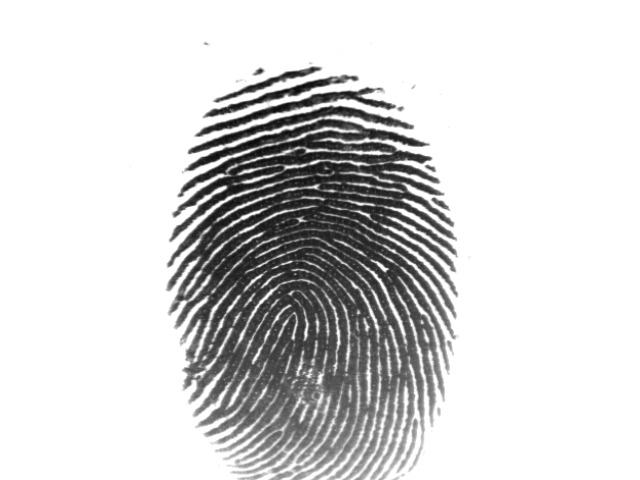}
		\label{nw}
	}		
	\subfigure[]
	{
		\includegraphics[width=0.9in]{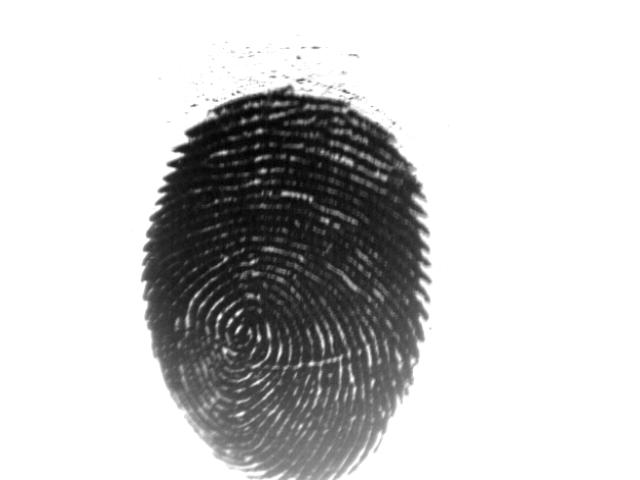}
		\label{wet}
	}
	
	\caption{Fingerprint images of different quality from FVC2004 $DB1$ database. (a) Dry (b) Normal dry (c) Good (d) Normal wet (e) Wet }
	\label{sampleimages}
\end{figure*}

In FQA, relevant quality features for each of the fingerprint images are computed. Most discriminative feature set is selected from all the extracted features which is fed as input to fuzzy c-means clustering. The fingerprint images are divided into five quality classes $Q_c$ where $c$ can be dry, normal dry, good, normal wet and wet. Properties of these defined quality classes are as follows:

\begin{itemize}
	
	\item \textbf{Dry} images are formed because of low pressure on the scanner surface or dryness of skin which produces broken ridges \cite{Yun2006101,MUNIR2012} as shown in Figure \ref{dry}. Large number of false minutiae can be detected in these images because of broken ridges.
	
	\item \textbf{Normal dry} quality fingerprint images contain scratchy ridges whose thickness is less than the subsequent valley region as shown in Figure \ref{nd}. In these images, most of the region has medium ridge-valley contrast \cite{MUNIR2012} and contains sufficient number of minutiae points for accurate recognition.
	
	\item \textbf{Good} quality fingerprint images are those which have clearly separated  ridge-valley structure \cite{Yun2006101,MUNIR2012} such that a minutiae extraction algorithm is able to operate very well as shown in Figure \ref{good}. These images have precisely located minutiae points which helps in improving recognition performance of the matching system.
	
	\item \textbf{Normal wet} fingerprint images as shown in Figure \ref{nw} have dark and hazy ridges with less valley region. These images contains sufficient number of minutiae points \cite{MUNIR2012} for accurate recognition.
	
	\item \textbf{Wet} fingerprint images are too dark and doesn't have clear ridge-valley separation which makes it quite difficult to separate ridge-valley structure in them. In these images, most of the valley region is filled with moisture due to high pressure or oily skin \cite{Yun2006101,MUNIR2012} as shown in Figure \ref{wet}. 
	
\end{itemize}


\subsubsection{Feature extraction}

Quality of a fingerprint is usually measured in terms of clarity of ridge and valley structure. Different states of this structure form different quality classes of fingerprints. These different classes of quality are caused by many factors such as user-specific factors like age, social customs, gender, and injuries or environmental issues such as temperature, humidity etc or user-sensor interaction based on indoor or outdoor use of sensors or dust on sensors etc. Therefore, to classify the fingerprint image in different quality classes some relevant features which exhibit properties of the images in those quality classes are extracted. We have extracted features like mean, variance, moisture, ridge valley area uniformity and ridge line count of a fingerprint image. Local variance based segmentation approach \cite{Mehtre1993} is followed to find the foreground region of the fingerprint image. All of these features are computed in a $16\times16$ block-wise manner in region of interest of the fingerprint image. We select a block of $16\times16$ to accommodate at least one ridge valley pair block since a ridge valley pair is 8-12 pixel wide in a 500 dpi fingerprint image \cite{maltoni2009handbook}. We extract the following five features: moisture, mean, variance, RVAU and RLC from the fingerprint images where moisture, RVAU and RLC are computed from the binary image obtained using Otsu’s method \cite{otsu1979}. Feature evaluation for the foreground blocks ($FB$) of fingerprint image is done as follows:


\begin{enumerate}[label=(\roman*)]
	
	\item \bf{Moisture (MI):} \normalfont Moisture level on a fingerprint will influence dry or wet nature of the fingerprint images. Therefore, moisture of a fingerprint image should be computed effectively to assess quality of a fingerprint. Moisture level of a block is evaluated as the percentage of unwanted ridge pixels in the valley region of a fingerprint block. The percentage of unwanted ridge pixels is obtained by subtracting percentage of average number of ridge pixels ($\lambda$) in a manually labeled good quality block from the percentage of total number of ridge pixels in that block. The value of $\lambda$ is obtained by averaging percentage of ridge (black) pixels in 500 manually labeled good quality blocks. Averaging the moisture level of all foreground blocks of a fingerprint image will give overall moisture of the fingerprint image $I$. Dry fingerprint images contain moisture level in negative, which indicates low moisture level on the fingertip. This low moisture level causes less contact with sensor platen, leading to broken ridges. On the contrary, positive moisture level of wet fingerprints indicates that some of the valley regions are filled with dark pixels. Computation of moisture ($MI^{local}_{FB}$) for $m \times n$ size foreground block $FB$ is performed using Eq. (\ref{MIlocal}) as below:	
	\begin{equation}
		MI^{local}_{FB}=\bigg(\frac{1}{m \times n}\sum\limits_{i=1}^{i=m}\sum\limits_{j=1}^{j=n}FB(i,j) \hspace*{0.3mm} [FB(i,j)=0]\bigg)\times100 - \lambda      
		\label{MIlocal}
	\end{equation}	
	\begin{equation}
\lambda=\Bigg\{\frac{\sum\limits_{{c=1}}^{{c=500}} \Big(\frac{1}{m \times n}\sum\limits_{i=1}^{i=m}\sum\limits_{j=1}^{j=n}B_{{c}_{good}}(i,j) \\\hspace*{0.3mm} [B_{{c}_{good}}(i,j)=0]\Big)}{500}\Bigg\}\times 100
	\label{lambda}
	\end{equation}		
	where, $MI^{local}_{FB}$ represents the moisture level of foreground block $FB$,  $FB(i,j)$ is the value at pixel location (i, j) of a binary block $FB$, $m \times n$ is the count of all pixels in the block ($16^2=256$) and $\lambda$ value is obtained as 51.25\% using Eq. (\ref{lambda})  where $B_{c_{good}}$ is manually labeled good quality block. Global moisture level ($MI^{global}$) of the fingerprint image is computed using Eq. (\ref{MIglobal}) where $|FB|$ is the count of number of foreground blocks in a fingerprint image.	
	\begin{equation}
MI^{global}=  \frac{1}{|FB|}\sum\limits_{i=1,j=1}^{i<=X,j<=Y} \hspace{-1mm} MI^{local}_{FB(i,j)}
	\label{MIglobal}
	\end{equation}		
	where, $MI^{local}_{FB(i,j)}$ is the moisture level of a foreground block and i and j is the horizontal and vertical index of $16 \times 16$ size blocks.	
	
	\item \bf{Mean (M): }\normalfont  Mean of the fingerprint image is considered only for the foreground area which show the overall gray level of the image. Different quality of fingerprint images have different mean distribution which makes it a good feature for quality prediction of fingerprint images. Mean ($M^{local}_{FB}$) of each foreground block $FB$ of size $m \times n$ in a fingerprint image is computed using Eq. (\ref{mean}).
	%
	\begin{equation}
		M^{local}_{FB}=\frac{1}{m\times n}\sum_{i=1}^{i=m}\sum_{j=1}^{j=n}FB(i,j)
	\label{mean}
	\end{equation}	
	global mean ($M^{global}$) of the fingerprint image is computed using Eq. (\ref{Mglobal}) where $|FB|$ is the count of number of foreground blocks in a fingerprint image.	
	\begin{equation}
		M^{global}=  \frac{1}{|FB|}\sum\limits_{i=1,j=1}^{i<=X,j<=Y} \hspace{-1mm}M^{local}_{FB(i,j)}		
	\label{Mglobal}
	\end{equation}	
	\item \bf{Variance (V): } \normalfont In order to obtain overall uniformity of gray level in the foreground region of the fingerprint image variance is calculated. Variance ($V^{local}_{FB}$) of each foreground block $FB$ is computed as given in Eq. ($\ref{varlocal}$). Overall variance ($V^{global}$) of fingerprint image is computed using Eq. ($\ref{varglobal}$)
	
	\begin{equation}
		V^{local}_{FB}=\frac{1}{m\times n}\sum\limits_{i=1}^{i=m}\sum\limits_{j=1}^{j=n}(FB(i,j)-M^{local}_{FB})^2		
	\label{varlocal}
	\end{equation}
	\begin{equation}
		V^{global}=  \frac{1}{|FB|}\sum\limits_{i=1,j=1}^{i<=X,j<=Y}\hspace{-4mm} V^{local}_{FB(i,j)}			
	\label{varglobal}
	\end{equation}	
	\item \bf{Ridge valley area uniformity (RVAU): } \normalfont In general, a good quality fingerprint image contains equally classified ridge-valley pattern with identical thickness throughout the foreground region of fingerprint image. But, it is found that the ridge-valley pattern across different foreground blocks in a fingerprint image is not uniform in real scenarios. Therefore, determining RVAU in a fingerprint image will help to predict the overall quality of the fingerprint image. Ratio of ridge area versus valley area is computed using total number of ridge and valley pixels in the binary image. RVAU ($RVAU^{local}_{FB}$) of a foreground block $FB$ is evaluated as defined in Eq. (\ref{RVAUlocal}).	
	\begin{equation}
		RVAU^{local}_{FB} = \frac{\sum_{i=1}^{i=m}\sum\limits_{j=1}^{j=n}B(i,j)[FB(i,j)=0]}{|m\times n|- \sum\limits_{i=1}^{i=m}\sum\limits_{j=1}^{j=m}FB(i,j)[FB(i,j)=0]}			
	\label{RVAUlocal}
	\end{equation}	
	global RVAU ($RVAU^{global}$) of a fingerprint image is computed using Eq. (\ref{RVAUglobal}) where $|FB|$ is the count of number of foreground blocks in a fingerprint image.		
	\begin{equation}
		RVAU^{global}=  \frac{1}{|FB|}\sum\limits_{i=1,j=1}^{i<=X,j<=Y} \hspace{-2mm} RVAU^{local}_{FB(i,j)}			
	\label{RVAUglobal}
	\end{equation}	
	%
	%
	\item \bf{Ridge line count (RLC): } \normalfont Counting the number of ridge lines in each foreground block provides some indication of the block quality which constitutes to overall quality of the fingerprint image. The motivation for considering RLC as a candidate feature to predict fingerprint quality is that a good fingerprint contains constant number of ridge lines for all the foreground blocks. It is  possible that some of the foreground blocks may have zero number of ridge lines which is possible in a dry image due to less contact with sensor platen. Similarly, wet fingers exhibit opposite effect where the sample looks very dark and some of the image blocks may contain only 1 ridge of abnormal thickness. Therefore, evaluation of RLC is required as it is a good indicator to determine overall fingerprint image quality.
	
	RLC for foreground block of a fingerprint image is computed using the ridge map of the block with one-pixel thickness. The ridge map is obtained by thinning morphological operation which is first rotated vertically using orientation estimation method adapted from \cite{tabassi05} to make the ridge lines vertical. The orientation of each block is computed using \cite{tabassi05}, where the numerical gradient of the block is determined using the finite central difference for all interior pixels in the x-direction ($dx$) and y-direction ($dy$). The covariance matrix ($ C_{FB}$) of the block using principal components analysis is computed using the following expression given in \cite{tabassi05}.	
	\begin{center}
		$ C_{FB}$=$\frac{1}{m*n}\sum_{m*n}^{}\begin{Bmatrix}
			\begin{bmatrix}
			dx \\
			dy
			\end{bmatrix}
			\begin{bmatrix}
			dx & dy
			\end{bmatrix}
			\end{Bmatrix}$=$\begin{bmatrix}
			a & c\\
			c & b
			\end{bmatrix}$
	\end{center}
	\vspace{2mm}
	orientation of a foreground block ($theta_{FB}$) is computed using following equation as given in \cite{tabassi05}.
	\begin{center}
		$theta_{FB}$= $\tan ^{ - 1} (\frac{c}{\sqrt{c^2+(a-b)^2}}\frac{a-b}{\sqrt{c^2+(a-b)^2}})$
	\end{center}
		Next, the block is rotated using the orientation of the block in such a way that the ridges are at the vertical position. Further, thinning morphological operation is performed to count the pixel bit-flips (white to black and black to white) for each row of a block. This gives the count of twice the number of ridges present in each row of a block. The half of the maximum of these counts is considered as RLC of the block ($RLC^{local}_{FB}$) which is defined in Eq. (\ref{RLClocal}).	
	\begin{equation}
	\begin{aligned}
		RLC^{local}_{FB}= \frac{1}{2} \times Max_i \bigg(\sum\limits_{j=1}^{j=n}FB(i,j)[b->w||w->b]\bigg)
	\hspace{1mm} for\hspace{1mm}  i = 1  \hspace{1mm}  to \hspace{1mm}  m	
	\end{aligned}
	\label{RLClocal}	
	\end{equation}	
	average RLC for the entire fingerprint image ($RLC^{global}$) is computed using Eq. (\ref{RLCglobal}).	
	\begin{equation}
		RLC^{global}=  \frac{1}{|FB|}\sum\limits_{i=1,j=1}^{i<=X,j<=Y}\hspace{-4mm} RLC^{local}_{FB(i,j)}		
	\label{RLCglobal}
	\end{equation}
	\item \bf{Other features: }\normalfont Apart from the proposed features, fingerprint quality is influenced by some of the other features defined in literature. Therefore, to make fingerprint quality clustering more robust few other features are also evaluated. In our proposed work, we are clustering fingerprint images in dry, normal dry, good, normal wet and wet classes. Therefore, features which influence these quality classes must be considered as a quality predictor of fingerprint images. Hence, two features namely, uniformity and contrast from \cite{nfiq2.0} and four features namely, radial power spectrum (RPS), ridge valley thickness uniformity (RVU), gabor and gabor-shen from \cite{olsen2016} are considered in the candidate feature vector for quality prediction of fingerprint images.

	%
	Finally, we have 11-dimensional feature vector ($F$) which affects our defined quality classes. The feature vector comprises 2 features of NFIQ 2.0 \cite{nfiq2.0}, 4 features
	from Olsen’s method \cite{olsen2016} and 5 proposed features evaluated in this paper. These features are: moisture, mean, variance, RVAU, RLC, uniformity, contrast, RPS, RVU, gabor and gabor-shen. This feature vector is fed as input to feature selection unit to select most discriminative feature subset for quality clustering of fingerprint images.
\end{enumerate}	
\begin{table}[]
	\centering
		\caption{Probability value ($p$) for individual feature ANOVA test}
	\label{anovaprob}
	  \resizebox{0.5\textwidth}{!}{
	
	\begin{tabular}{|c|c|}
		\hline
		\multicolumn{1}{|c|}{\textbf{Feature}}                    & \multicolumn{1}{c|}{\textbf{Probability Value (p)}} \\ \hline
		\multicolumn{1}{|c|}{Uniformity}                      & \multicolumn{1}{c|}{1.38 e-41}          \\ \hline
		\multicolumn{1}{|c|}{Contrast}                        & \multicolumn{1}{c|}{4.49 e-71}          \\ \hline
		\multicolumn{1}{|c|}{Mean}                            & \multicolumn{1}{c|}{1.50 e-78}          \\ \hline
		\multicolumn{1}{|c|}{Moisture}                       & \multicolumn{1}{c|}{7.40 e-76}          \\ \hline
		\multicolumn{1}{|c|}{Variance}             & \multicolumn{1}{c|}{6.23 e-60}          \\ \hline
		\multicolumn{1}{|c|}{Ridge line count}              & \multicolumn{1}{c|}{2.54 e-14}          \\ \hline
		\multicolumn{1}{|c|}{Ridge valley area uniformity} & \multicolumn{1}{c|}{5.93 e-42}          \\ \hline
		\multicolumn{1}{|c|}{Radial power spectrum}         & \multicolumn{1}{c|}{1.38 e-07}          \\ \hline
		\multicolumn{1}{|c|}{Ridge valley Uniformity}       & \multicolumn{1}{c|}{1.48 e-43}          \\ \hline
		\multicolumn{1}{|c|}{Gabor}                           & \multicolumn{1}{c|}{2.61 e-15}          \\ \hline
		\multicolumn{1}	{|c|}{Gabor-shen}                                          & \multicolumn{1}	{|c|}{2.88 e-15}                               \\ \hline 
	
	\end{tabular}}

\end{table}

\begin{table}[b]
	\centering
	\caption{Probability value ($p$) corresponding to each possible pair of dry (D), wet (W), good (G), normal dry (ND) and normal wet (NW)  cluster for individual features}		
	\label{tukeyprob}
	  \resizebox{\textwidth}{!}{
	\begin{tabular}{|c|c|c|c|c|c|c|c|c|c|c|}
		\hline
		\textbf{Feature}                    & \textbf{(D, W)} & \textbf{(D, G)} & \textbf{(D, ND)} & \textbf{(D, NW)} & \textbf{(W, G)} & \textbf{(W, ND)} & \textbf{(W, NW)} & \textbf{(G, ND)} & \textbf{(G, NW)} & \textbf{(ND, NW)} \\ \hline
		Uniformity                     & 0.0000            & 0.0000            & 0.0000             & 0.0000             & 0.0000            & 0.0000             & 0.0000             & 0.0128             & 0.0031             & 0.0000              \\ \hline
		Contrast                       & 0.0000            & 0.0000            & 0.0000             & 0.0000             & 0.0000            & 0.0000             & 0.0000             & 0.0016             & 0.0000             & 0.0000              \\ \hline
		Mean                            & 0.0000            & 0.0000            & 0.0000             & 0.0000             & 0.0000            & 0.0000             & 0.0000             & 0.0000             & 0.0000             & 0.0000              \\ \hline
		Moisture                       & 0.0000            & 0.0000            & 0.0000             & 0.0000             & 0.0000            & 0.0000             & 0.0000             & 0.0000             & 0.0000             & 0.0000              \\ \hline
		Variance             & 0.0000            & 0.0000            & 0.0000             & 0.0000             & 0.0000            & 0.0000             & 0.0000             & 0.0000             & 0.0000             & 0.0000              \\ \hline
		Ridge line count              & 0.0012            & 0.0002            & 0.0048             & 0.0282             & 0.0000            & 0.0000             & 0.0000             & \textbf{0.9182}    & \textbf{0.6188}    & \textbf{0.9781}     \\ \hline
		Ridge valley area uniformity & 0.0000            & 0.0000            & 0.0456             & 0.0000             & 0.0000            & 0.0000             & 0.0000             & 0.0473             & 0.0463             & 0.0000              \\ \hline
		Radial power spectrum         & 0.020             & 0.0000            & 0.0000             & 0.0001             & 0.0482            & \textbf{0.4460}    & \textbf{0.5984}    & \textbf{0.8233}    & \textbf{0.6883}    & \textbf{0.9993}     \\ \hline
		Ridge valley uniformity       & 0.0000            & 0.0000            & 0.0117             & 0.0000             & 0.0000            & 0.0000             & 0.0000             & 0.0005             & 0.0027             & 0.0000              \\ \hline
		Gabor                           & 0.0000            & 0.0000            & 0.0000             & 0.0000             & 0.0196            & \textbf{1.0000}    & 0.0195             & 0.0173             & \textbf{1.0000}    & 0.0172              \\ \hline
		Gabor-shen                     & \textbf{0.9923}   & 0.0000            & 0.0000             & 0.0000             & 0.0000            & 0.0000             & 0.0000             & \textbf{0.3963}    & \textbf{0.6619}    & \textbf{0.9933}     \\ \hline
	\end{tabular}}

\end{table}

\subsubsection{Feature selection}

\begin{figure}[p]
	\centering	
	
	\subfigure[]
	{
		
		\includegraphics[height=2.2cm,width=6.5cm]{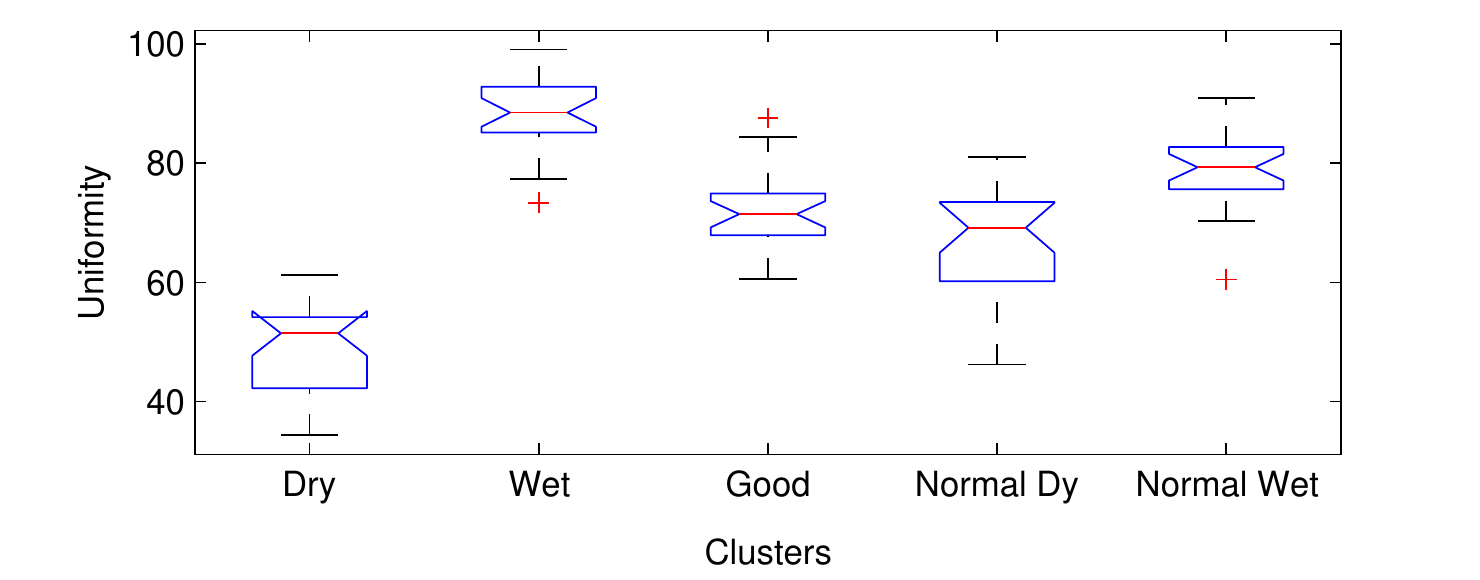}

	}
 \vspace{-2mm}
	\subfigure[]
	{
	\includegraphics[height=2.2cm,width=6.5cm]{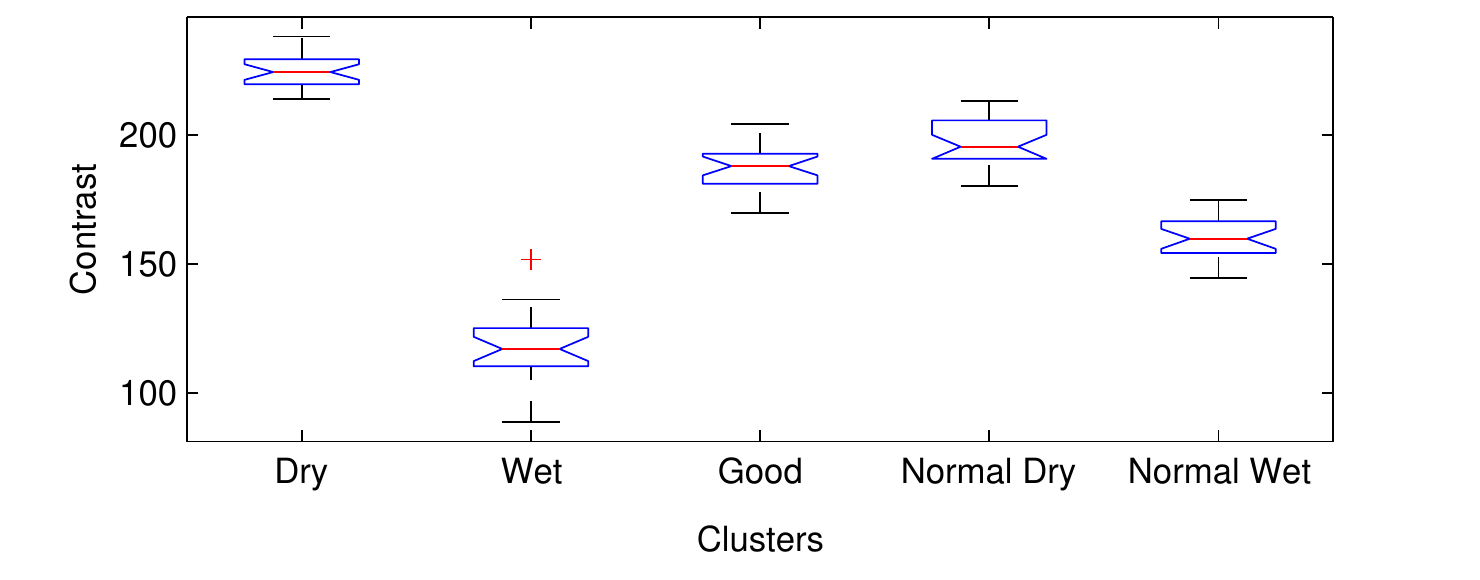}

	}

	\subfigure[]
	{
		\includegraphics[height=2.2cm,width=6.5cm]{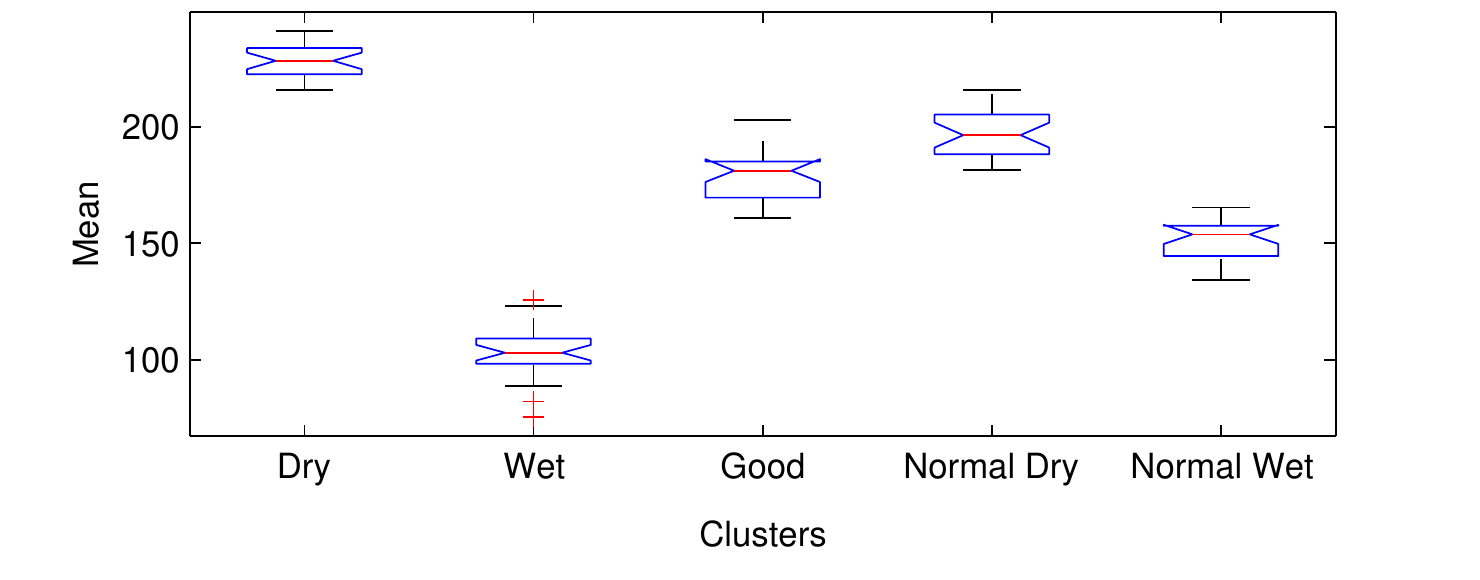}

	}
\vspace{-2mm}
	\subfigure[]
	{
		\includegraphics[height=2.2cm,width=6.5cm]{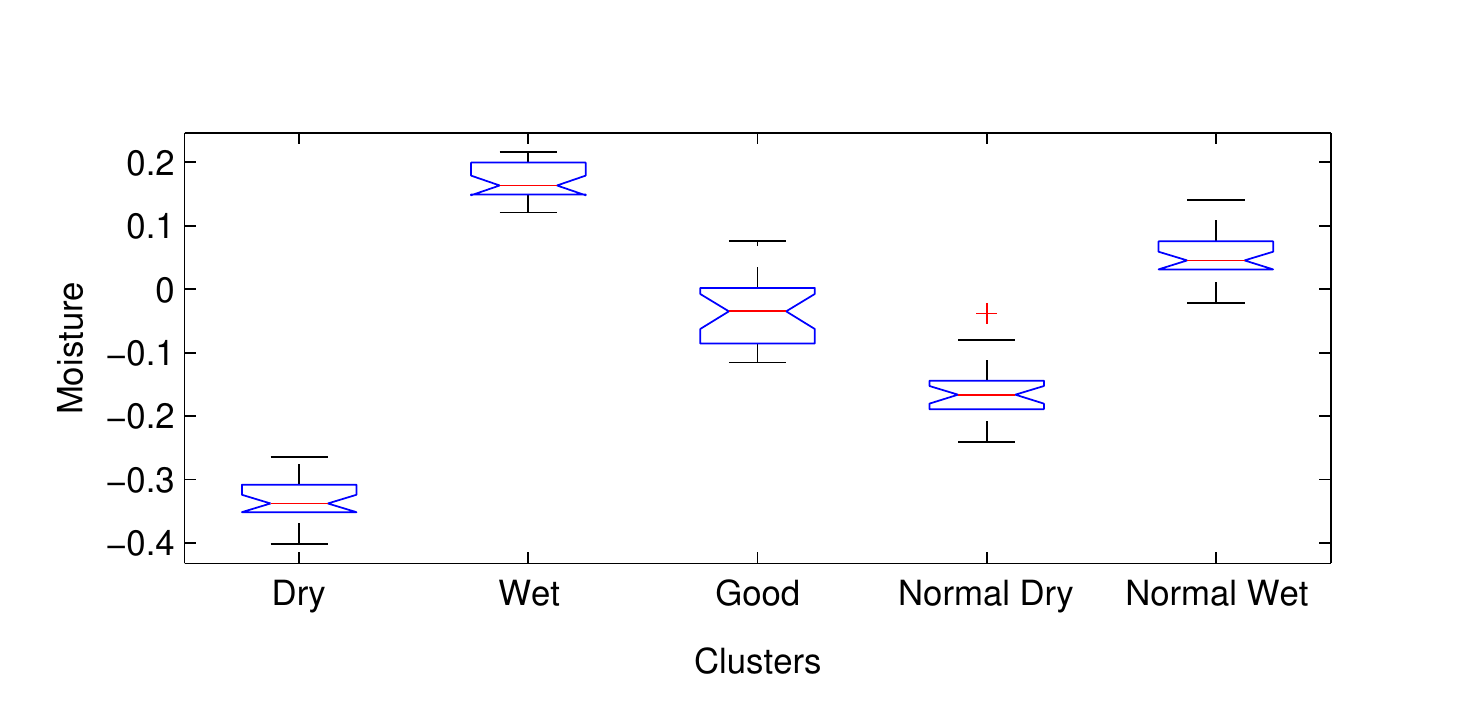}
		\label{fig:second_sub}
	}	
	\subfigure[]
	{
		\includegraphics[height=2.2cm,width=6.4cm]{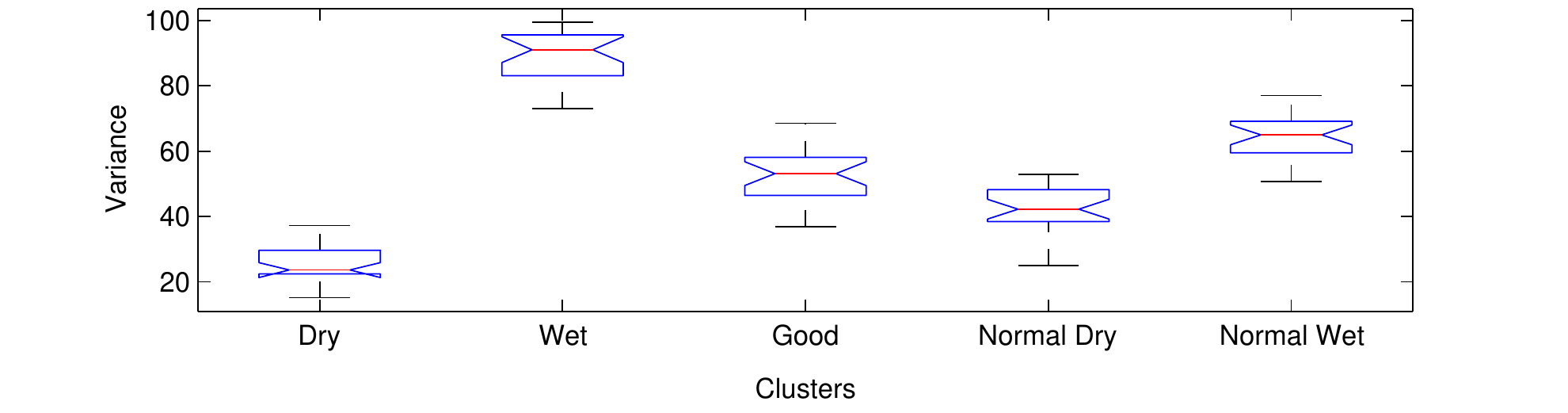}
		\label{fig:second_sub}
	}
\vspace{-2mm}
	\subfigure[]
	{
		\includegraphics[height=2.2cm,width=6.5cm]{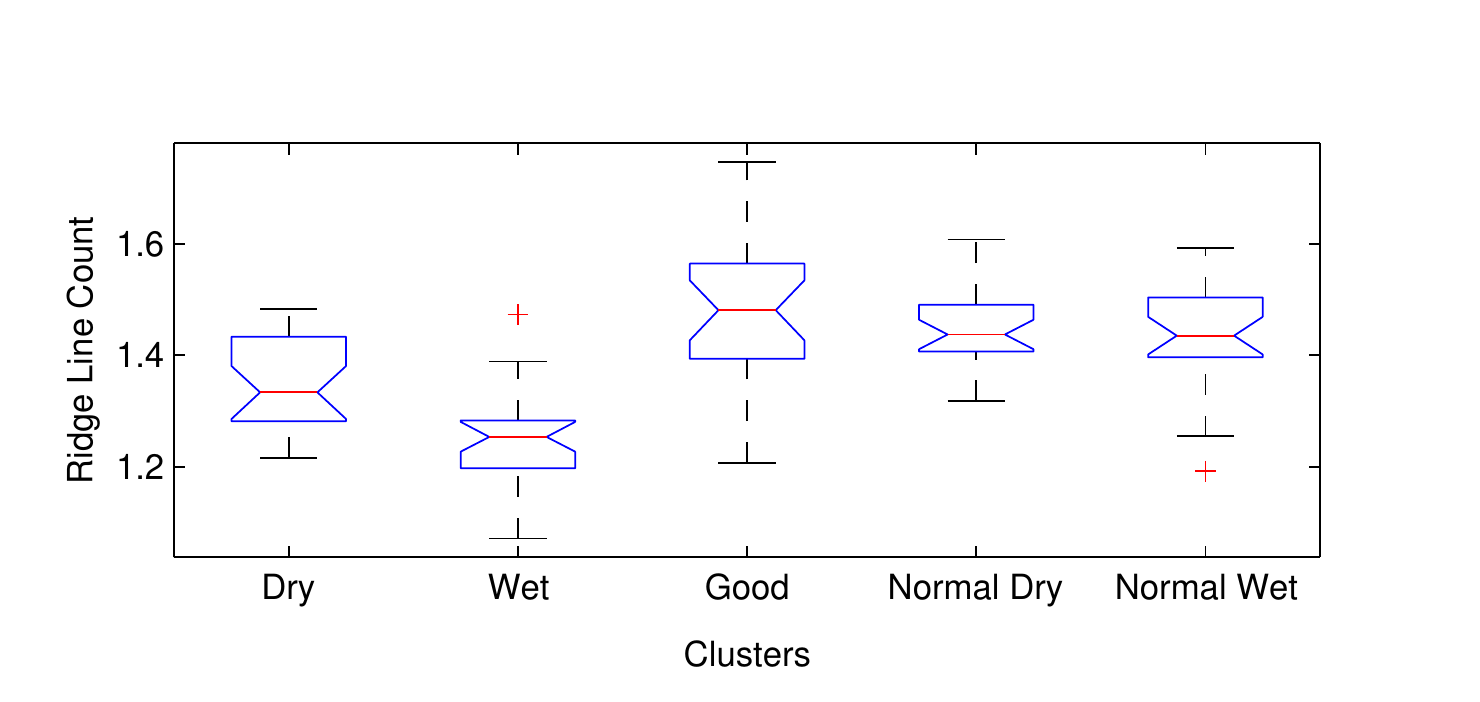}
		\label{fig:second_sub}
	}

		\hspace{4mm}
	
	\subfigure[]
	{
		\includegraphics[height=2.2cm,width=6.4cm]{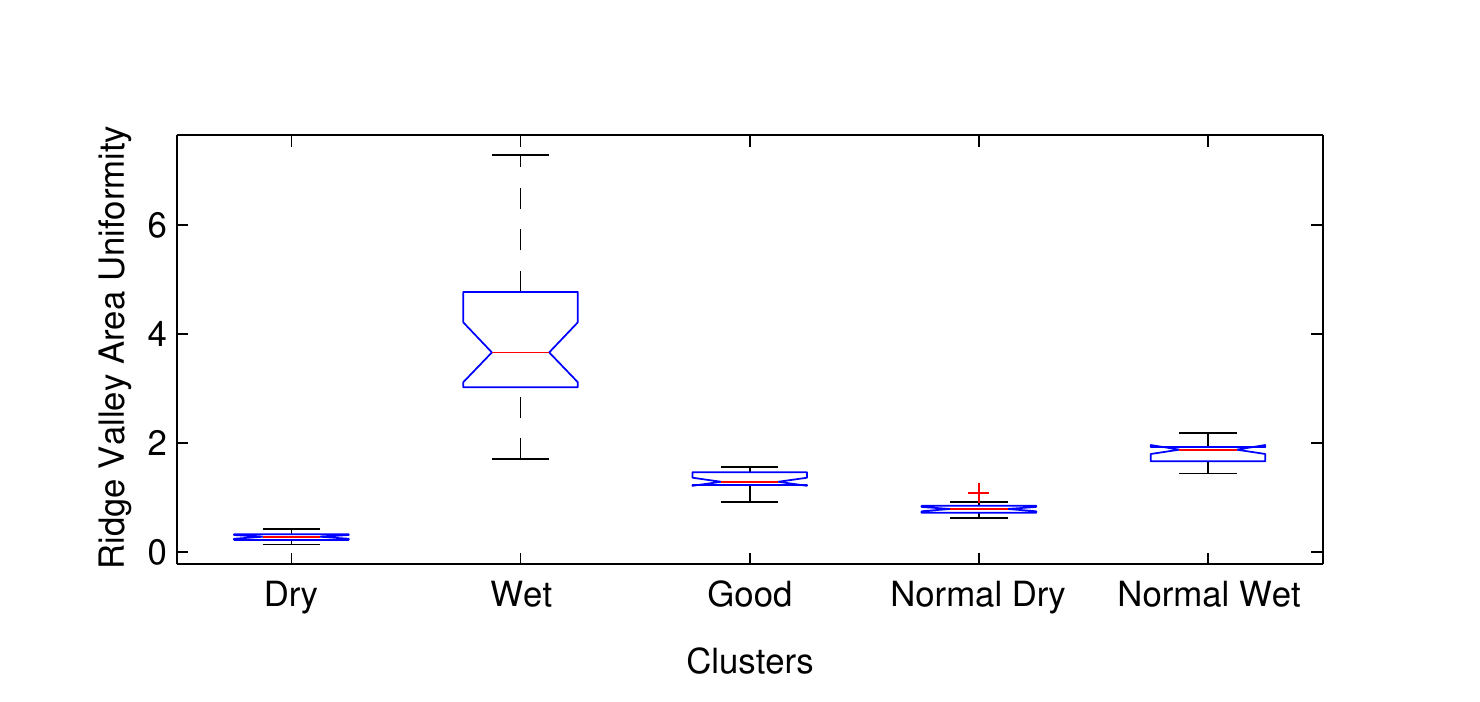}
		\label{fig:second_sub}
	}	
\vspace{-2mm}	
	\subfigure[]
	{
		\includegraphics[height=2.2cm,width=6.5cm]{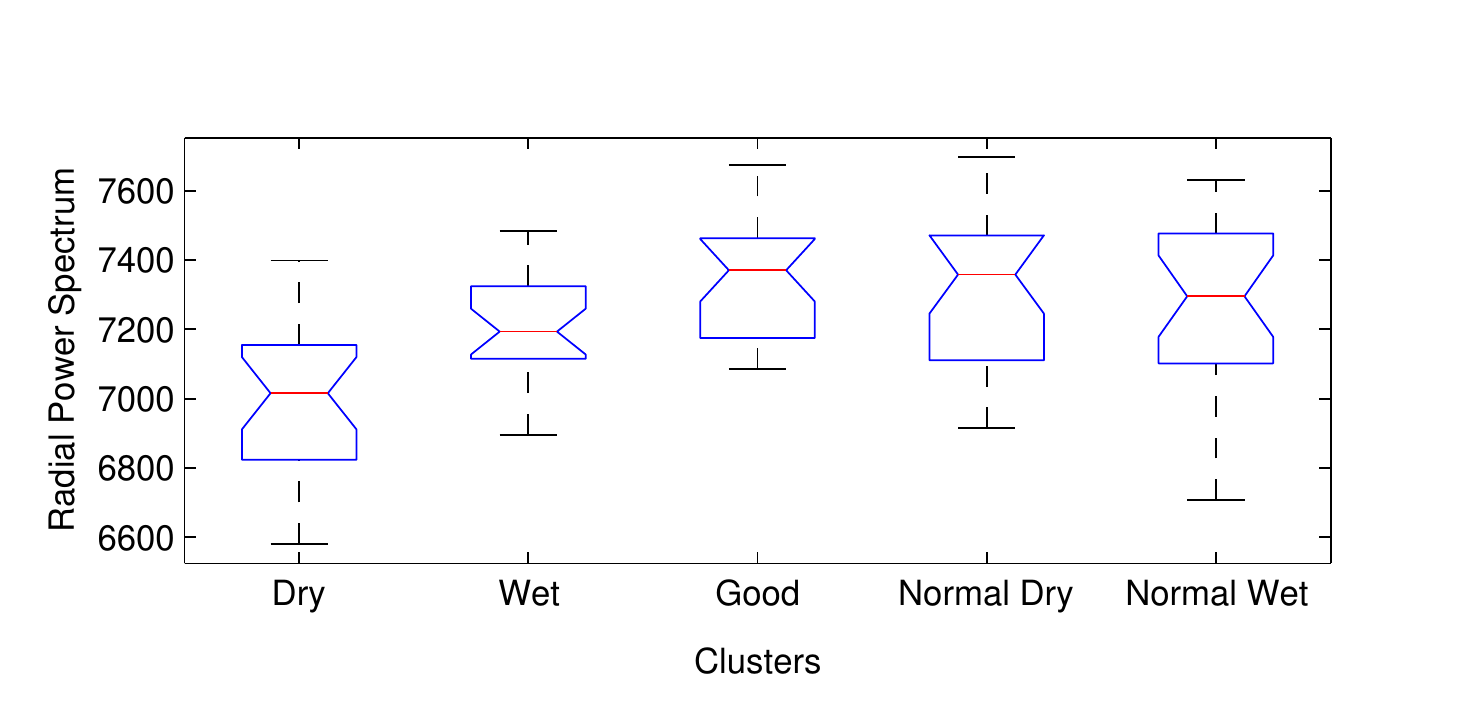}
		\label{fig:second_sub}
	}

	\subfigure[]
	{
		\includegraphics[height=2.2cm,width=6.5cm]{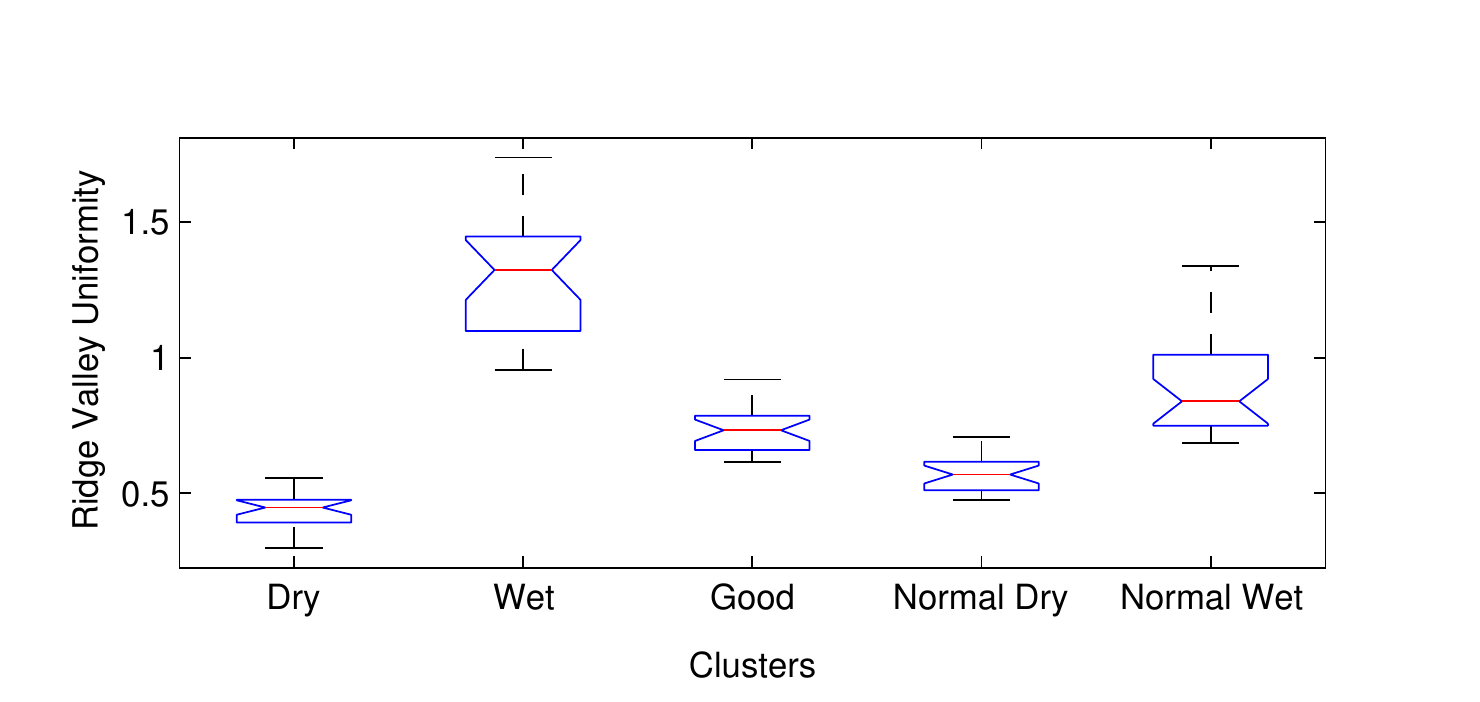}
		\label{fig:second_sub}
	}
     \vspace{-2mm}	
	\subfigure[]
	{
	\includegraphics[height=2.2cm,width=6.5cm]{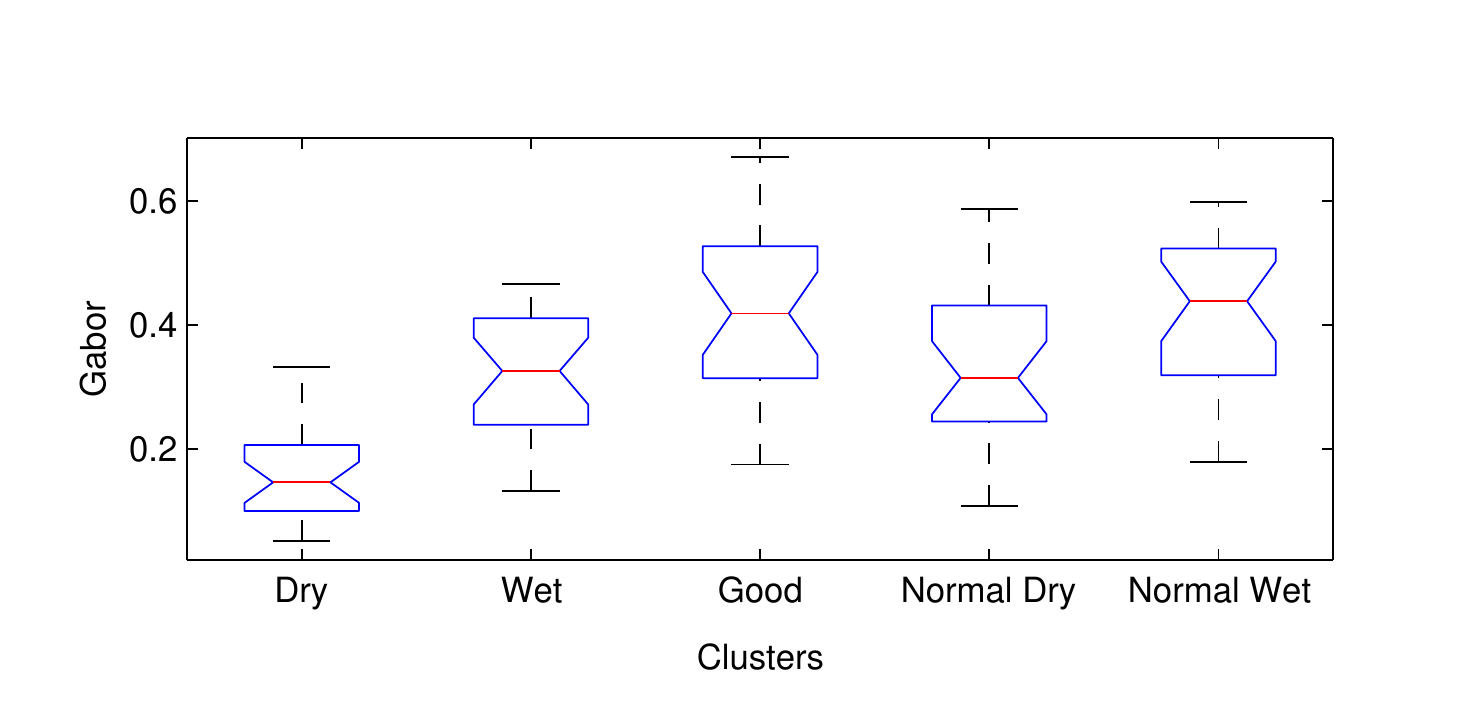}
		\label{fig:second_sub}
	}
	\vspace{-2mm}
	\subfigure[]
	{
\includegraphics[height=2.2cm,width=6.5cm]{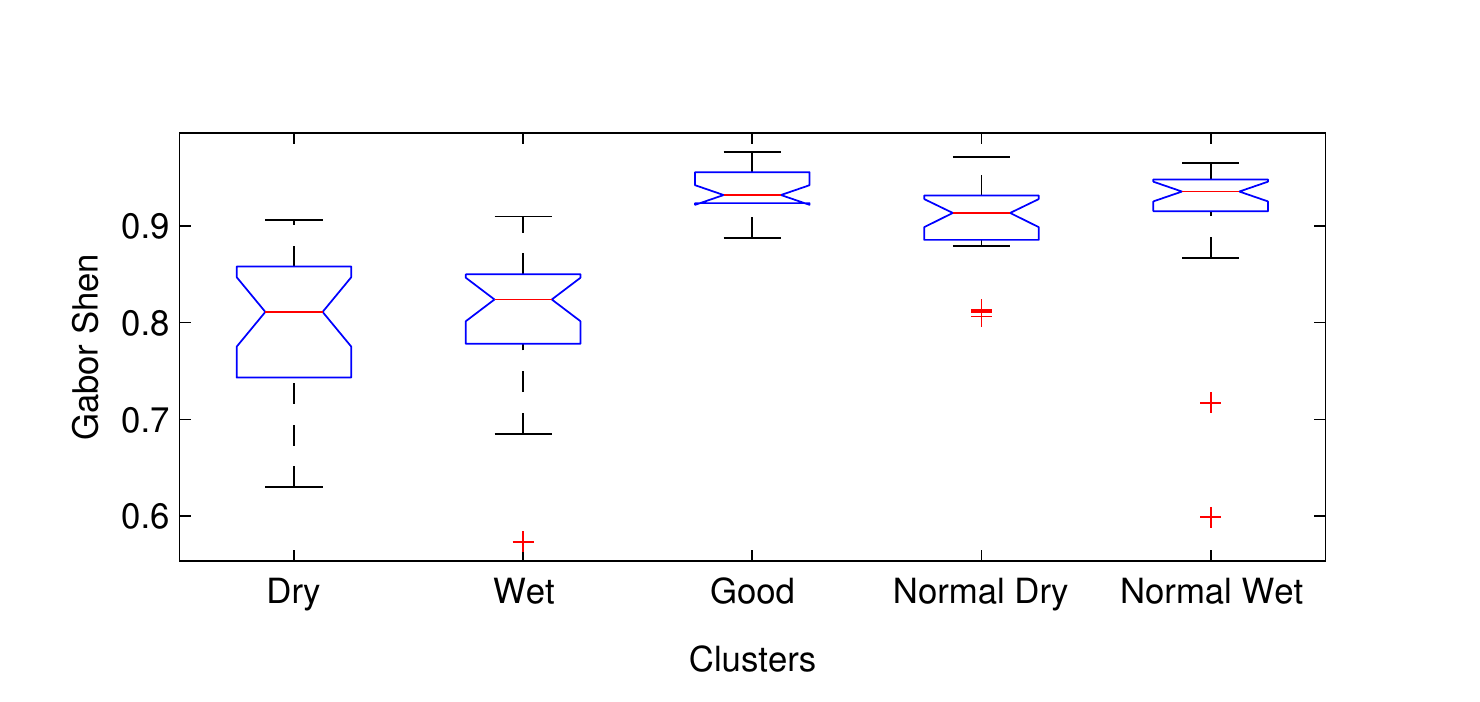}
		\label{fig:second_sub}
	}

	\caption{One way ANOVA test for individual features: (a) Uniformity (b) Contrast (c) Mean  (d) Moisture (e) Variance (f) Ridge line count (g) Ridge valley area uniformity (h) Radial power spectrum (i) Ridge valley uniformity (j) Gabor (k) Gabor-shen}
	\label{featureselection}
	
\end{figure}

Fingerprint images of all the four datasets DB1-DB4 of FVC2004 \cite{Maio2004} are clustered manually in the defined quality classes $Q_c$ (dry, normal dry, good, normal wet and wet). Feature vector $F$ from  200 fingerprint images of each quality class is constituted  to present as input to one-way ANOVA test. One way ANOVA computes the dissimilarity between the cluster and dissimilarity within the cluster to estimate that the defined classes are similar with probability $p$. A good feature exhibits more dissimilarity between the classes and less dissimilarity within the class. Therefore, to show the statistical significance of each feature, a null hypothesis is tested. Null hypothesis affirms that there is no significant difference between the means of the different classes for a particular feature ($f_1, f_2 ....f_{11}$). Acceptance or rejection of null hypothesis depends on whether the probability value $p$ is less or more than the significance level ($\alpha =0.05$). Results of one way ANOVA with probability value $p$ for the individual feature is shown in Table \ref{anovaprob}. The $p$-value obtained for every feature ($f_1, f_2 ....f_{11}$) is too less than $0.05$ therefore null hypothesis is rejected for all the features. Rejection of null hypothesis reflects that means of each feature for the defined quality classes are not same. This signifies that a considerable difference lies in all the 5 defined quality classes. Feature distribution obtained by one way ANOVA for each quality class is illustrated in Figure \ref{featureselection}.

One way ANOVA test proves that there is some statistical difference in all quality classes for each feature but it doesn't show where this difference lies or whether there is enough statistical difference in each pair of the quality classes. To gain more insight into this, Tukey's honest significant difference (HSD) test is performed. This test identifies whether there is significant difference between mean feature values for each pair (e.g. (D,ND) or (ND,G) etc.) of the quality classes. Probability value $p$ corresponding to each pair of the quality class for the individual feature is given in Table \ref{tukeyprob}. It can be observed that probability value $p$ for some pairs of quality bins is more than the $\alpha$ value ($p>0.05$). This indicates that mean feature value for the quality classes of these pairs are not significantly different from each other. Hence, these features are not well suited for quality clustering of fingerprint images and they can be removed from feature vector $F$. Probability value $p$ for (good-normal dry), (good, normal wet) and (normal dry, normal wet) pairs of ridge line count is 0.9182, 0.6188 and 0.9781, which is more than the significance level ($\alpha=0.05$). Therefore, null hypothesis is accepted for these pairs of RLC feature, stating that the mean RLC value for the fingerprint images in these pairs of classes are similar. It can be verified in Figure \ref{featureselection}(f) that RLC values for fingerprint images in these pairs of clusters fall in almost same range.  Therefore, RLC feature is insignificant for discriminating different quality fingerprint images and thus removed from $F$.
Similarly, significance level ($\alpha$) for radial power spectrum feature have $p$-value ($0.4460>\alpha$), ($0.5984>\alpha$), ($0.8233>\alpha$), ($0.6883>\alpha$) and ($0.9993>\alpha$) respectively corresponding to (wet-normal dry), (wet-normal wet), (good-normal dry), (good-normal wet), and (normal dry-normal wet) fingerprint quality classes respectively (See Figure \ref{featureselection}(h) for feature distribution). Therefore, RPS is removed from feature vector $F$. Two pairs of quality classes (wet,normal dry) and (good-normal wet) have $p$ value higher than $\alpha$ ($1.000>\alpha$) for gabor feature and four pairs of quality classes for gabor-shen feature namely, (dry,wet)-($0.9923>\alpha$), (good,normal dry)-($0.3963>\alpha$), (good,normal wet)-($0.6619>\alpha$) and (normal dry,normal wet)-($0.9933>\alpha$) have higher probability than significance level. Therefore, Gabor (See Figure \ref{featureselection}(j) for feature distribution) and Gabor-shen feature (See Figure \ref{featureselection}(k) for feature distribution ) are removed from feature vector $F$. Similar feature value distribution is the rational behind high probability $p$ between some pairs of quality cluster. None of the other feature have probability value $p$ larger than $\alpha=0.05$ for any pair of quality classes in Table \ref{tukeyprob}. Therefore, final feature vector $F_s$ after feature selection contains 7 features namely, uniformity (U), contrast (C), mean (M), moisture (MI), variance (V), ridge valley area uniformity (RVAU) and ridge valley uniformity (RVU) for quality clustering of fingerprint images.  	

Further, fuzzy c-means clustering is used to cluster fingerprint images of FVC2004 datsets \cite{Maio2004} based on the selected feature set $F_s$.

\subsubsection{Quality clustering using fuzzy c-means}
The fuzzy c-means clustering algorithm is used for quality clustering of fingerprint images. Fuzzy c-means classifies fingerprint images in quality clusters $Q_c$. Unlike k-means clustering where each image must exclusively belong to one cluster, n membership values (between 0-1) are assigned to each image corresponding to n cluster center in fuzzy c-means. More the image is closer to a cluster center, its membership value for that cluster increases. Hence, the image will be assigned to the cluster which have the maximum membership for that image. Summation of the membership values assigned to each image corresponding to $Q_c$ clusters is always 1.
Selected feature set $F_s$ is fed as input to the fuzzy c-means algorithm for quality clustering of fingerprint images. Each fingerprint image of FVC2004 dataset is assigned any of the defined quality clusters $Q_c$ depending on its feature vector $F_s$.

\subsection{Two-stage fingerprint quality enhancement}

The proposed fingerprint quality enhancement works in two stages. In the first stage, a quality adaptive preprocessing is performed which adapts its parameter based on the quality of the fingerprint images. The second stage of FQE utilizes Gabor, STFT and ODF enhancement algorithms to enhance the fingerprint images using there contextual (orientation and frequency etc.) information. Description of both stages is given in following section.

\subsubsection{Quality adaptive preprocessing}

An adaptive fingerprint preprocessing method with different image characteristics is better than uniform filtering for all the images. Conventional filtering of dry fingerprint images can result into removal of black pixels of ridges which are thinner than neighboring ridges. Similarly, removing black pixels in valley region of oily images using conventional filtering can eradicate valleys that are very thin. Adaptive filtering adjusts parameters of the filter according to characteristics (dry, wet and good) of the fingerprint image. QAP method is illustrated in Figure \ref{blockdiagram} as first stage of the proposed FQE scheme. QAP method processes the fingerprint image based on the its quality nature $Q_c$ before the second stage enhancement process. Detailed description of each processing block is given in the following sections:

\paragraph{Unsharp Masking}
The initial step of the preprocessing method utilize unsharp masking with quality adjustable parameter values for enhancing fingerprint images of each quality bin.
Unsharp masking filter enhances the brightness of ridges and makes them sharper in dry and normal dry fingerprint images. On the contrary, it clears the valley region of wet and normal wet images by adjusting its radius and amount parameter. Radius parameter controls the size of the region around edge pixel that is affected by sharpening while amount parameter controls how much darker or brighter the pixels will be made. As good quality fingerprint images already have bright ridge pixels and clear valley region so unsharp masking filter parameters for them should be in such a way that it should not degrade there quality. Experimentally it is found that radius=2 and amount=1 doesn't degrade good quality images additionally it helps in making valley region more clear for them. Optimal parameter values for dry, normal dry, wet and normal wet fingerprint images are obtained utilizing parameter values of good images. As dry fingerprint images have very thin and less bright ridges with clear valley region so the radius of unsharp mask filter should be less and amount should be more to make ridge pixel brighter. Conversely, in wet fingerprint images the major objective is to make valley region clear  consequently radius should be more and amount of unsharp effect should be less. In order to obtain these parameter values highest membership value ($m$) given to a fingerprint image by fuzzy c-means clustering is used. Along with the $m$ value, coefficients are decided for each quality class. The quality gap between dry and good fingerprint images is more with respect to normal dry and good images. Therefore, increase or decrease in parameter value for dry is more than normal dry images. Same is true for the wet and normal wet fingerprint images. The coefficients for dry and wet fingerprint images is $c_{dry \hspace{1mm}OR \hspace{1mm}wet}=1$ and for normal dry and normal wet fingerprint images it is $c_{nd \hspace{1mm}OR\hspace{1mm} nw}=0.5$. Computation of radius and amount parameter value for dry and normal dry fingerprint images is done using Eq. (\ref{Rdrynd}) and Eq. (\ref{Adrynd}) respectively.
\begin{equation}
	R_{quality} = R_{good} - m_{quality} \times c_{quality}
\label{Rdrynd}
\end{equation}
\begin{equation}
	A_{quality} = A_{good} + m_{quality} \times c_{quality}
\label{Adrynd}
\end{equation}
here, quality can be dry or normal dry based on fingerprint characteristics, $m_{quality}$ is membership value for dry or normal dry fingerprint image for their respective cluster. Value of m will be in the range of [0,1] and coefficient value for dry and normal dry is $c_{dry}=1$  and $c_{nd}=0.5$ respectively. Similarly, radius and amount parameter value for wet and normal wet fingerprint images are obtained from Eq. (\ref{Rwetnw}) and Eq. (\ref{Awetnw}) respectively. Coefficient value for wet and normal wet are $c_{wet}=1$ and $c_{nw}=0.5$.

\begin{equation}
	R_{quality} = R_{good} + m_{quality} \times c_{quality}
\label{Rwetnw}
\end{equation}

\begin{equation}
	A_{quality} = A_{good} - m_{quality} \times c_{quality}
\label{Awetnw} 
\end{equation}

\paragraph{Contrast Limited Adaptive Histogram Equalization}
After quality adaptive unsharp masking, the contrast of the fingerprint image is enhanced using CLAHE. CLAHE is more suitable than normal histogram equalization for improving the local contrast of the fingerprint images as amplification of noise is less in CLAHE with respect to normal histogram equalization. Contrast limited adaptive histogram equalization enhances the contrast of fingerprint image while preserving brightness which enhances definitions of edges in different regions of the fingerprint image.

\paragraph{Gaussion Smoothing}  
Last preprocessing step in the first stage enhancement of fingerprint images is Gaussian smoothing. Gaussian smoothing is computational efficient and gives higher significance to the pixels near the edge (ridges) which avoids blurring effect near the ridges. Therefore, Gaussian smoothing is appropriate to reduce the effect of noise caused by contrast enhancement using CLAHE.

\subsubsection{Second stage enhancement}
Enhanced fingerprint images obtained from quality adaptive preprocessing in first stage are fed to the second stage enhancement. Second stage enhancement employ well known Gabor, STFT and ODF enhancement algorithms to further enhance the fingerprint images.
	\vspace{-2mm}
\section{Experimental results}\label{Experiments}
\subsection{Database and experimental methodology}

To assess the efficacy of the proposed FQA and FQE method, experiments are conducted on very low and varying quality FVC2004 fingerprint database \cite{Maio2004}. The FVC2004 database is formed to give a challenging benchmark for state-of-the-art recognition algorithms than previous fingerprint verification competitions \cite{Cappeli2006}. FVC2004 database consists of four datasets, namely, $DB1$, $DB2$, $DB3$ and $DB4$. Acquisition of fingerprints in $DB1-DB3$ is done in different sessions with varying conditions to enforce challenging image quality characteristics for recognition while $DB4$ consists of synthetically generated fingerprint images. The  images of the first two datasets $DB1-DB2$ are acquired using an optical sensor, $DB3$ with thermal sweeping sensor and $DB4$ with SFinGe v3.0 sensor. Each dataset consists of 100 finger with 8 impressions which makes each dataset of 800 fingerprint images.

Fingerprint verification tests are performed as per FVC protocol to ensure comparability of the results with \cite{journals/pami/HongWJ98,Gottschlich2012,Chikkerur2007198}. In order to obtain genuine match rate, each of the 8 samples of a subject in $DB1-DB4$ are compared with each other. Hence, there are $8C_2\times 100=2800$ genuine comparisons. False match rate is obtained by comparing  first sample of a subject with the first sample of the remaining 99 subjects. Therefore, total imposter comparisons are $100C_2=4950$. Description of FVC protocol and computation of equal error rate (EER) is explained in \cite{MaioFVC2000}. The EER points to a system’s operating point at which it correctly recognizes genuine users and imposters with equal probability. National institute of standard and technology biometric image software package (NBIS \cite{NBIS2007}) is employed for fingerprint matching. Minutiae are extracted using MINDTCT package and templates are matched by BOZORTH3 package of NBIS. 
%

\subsection{Experimental results of FQA}
\begin{figure}[h]
		\centering

		\subfigure[]
		{
			
			\includegraphics[width=0.1\textwidth]{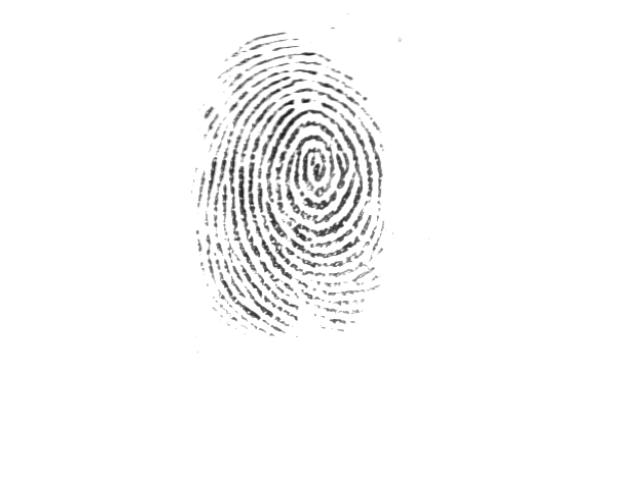} 
			\includegraphics[width=0.1\textwidth]{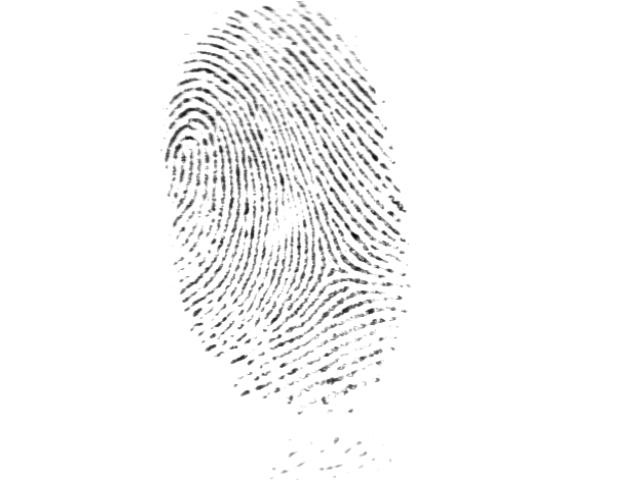} 
			\includegraphics[width=0.1\textwidth]{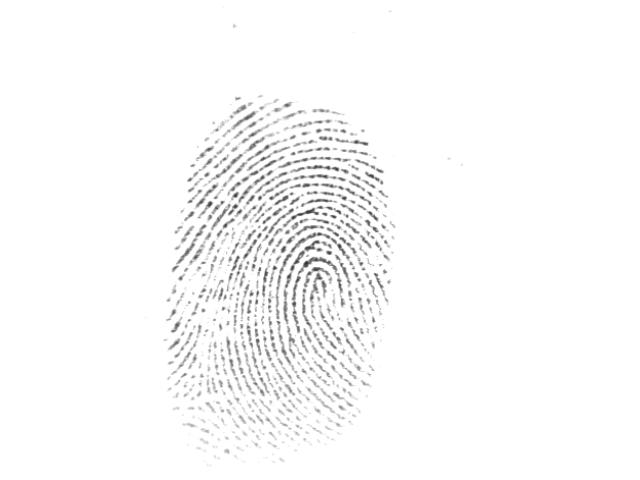} 
			\includegraphics[width=0.1\textwidth]{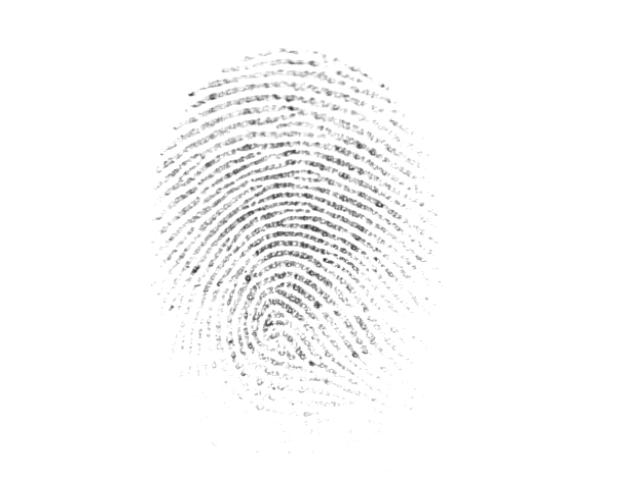} 
			\label{fig:second_sub}			
		}
		\hspace{2mm}
		\subfigure[]
		{
			
			\includegraphics[width=0.1\textwidth]{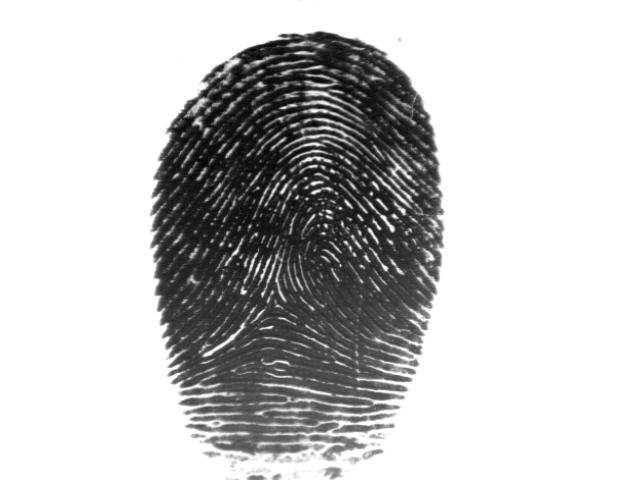} 
			\includegraphics[width=0.1\textwidth]{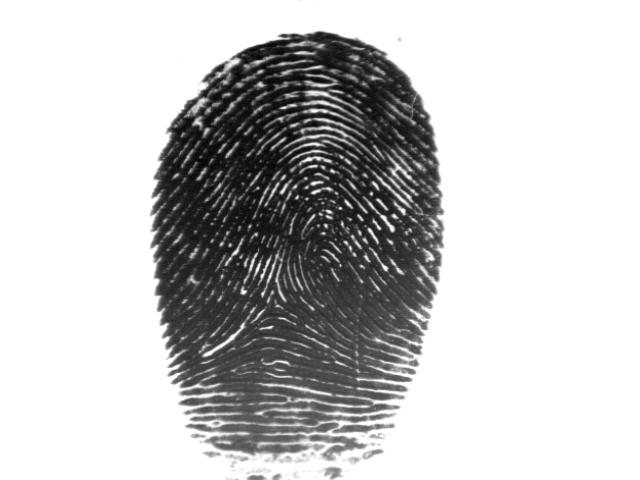} 
			\includegraphics[width=0.1\textwidth]{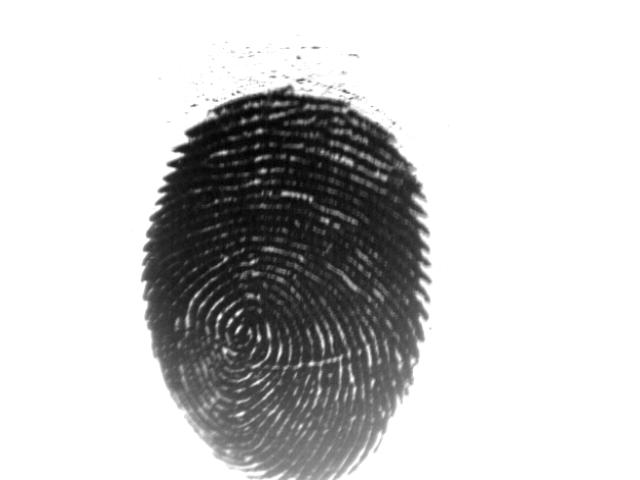} 
			\includegraphics[width=0.1\textwidth]{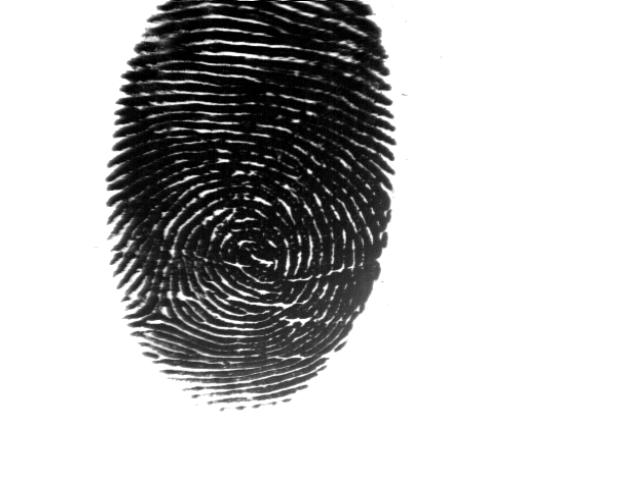} 
			\label{fig:second_sub}

		}
		\subfigure[]
		{
			
			\includegraphics[width=0.1\textwidth]{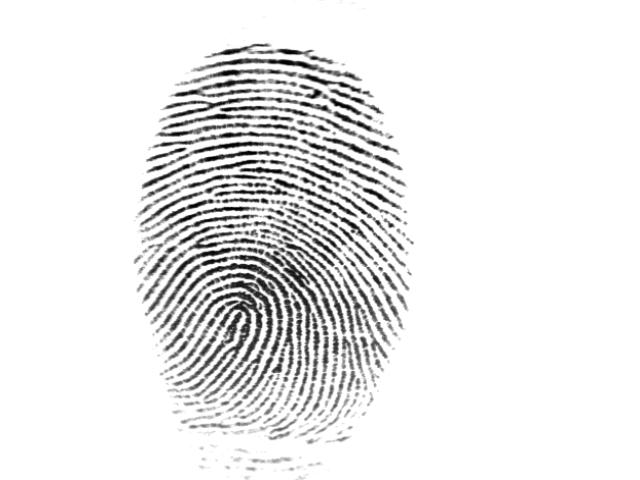} 
			\includegraphics[width=0.1\textwidth]{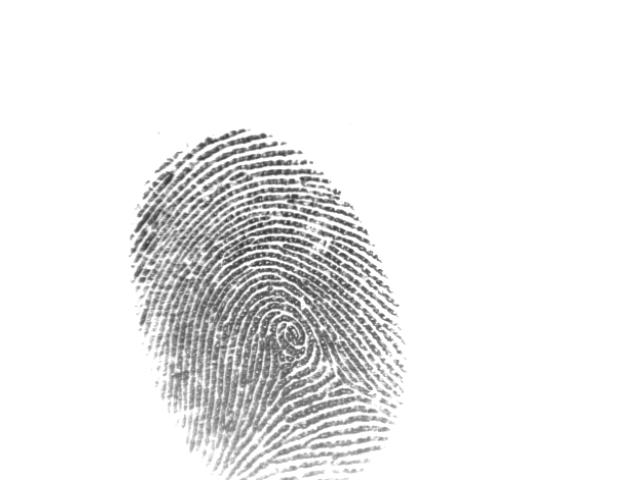} 
			\includegraphics[width=0.1\textwidth]{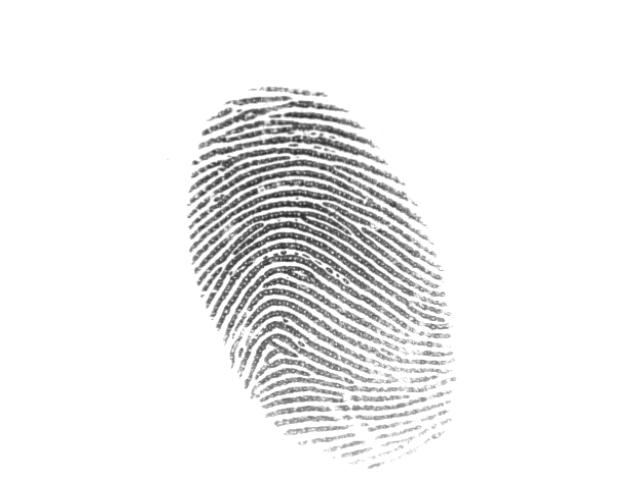} 
			\includegraphics[width=0.1\textwidth]{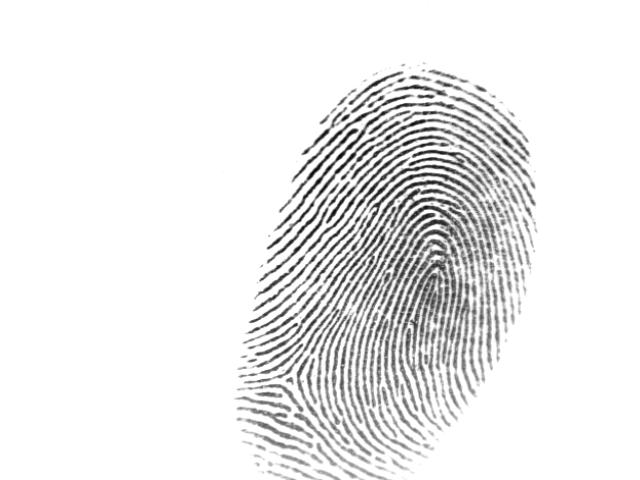} 
			\label{fig:second_sub}

		}	
		\hspace{2mm}		
		\subfigure[]
		{
			
			\includegraphics[width=0.1\textwidth]{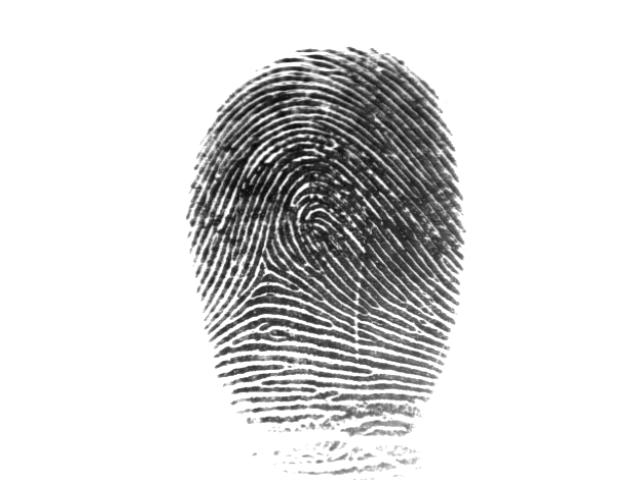} 
			\includegraphics[width=0.1\textwidth]{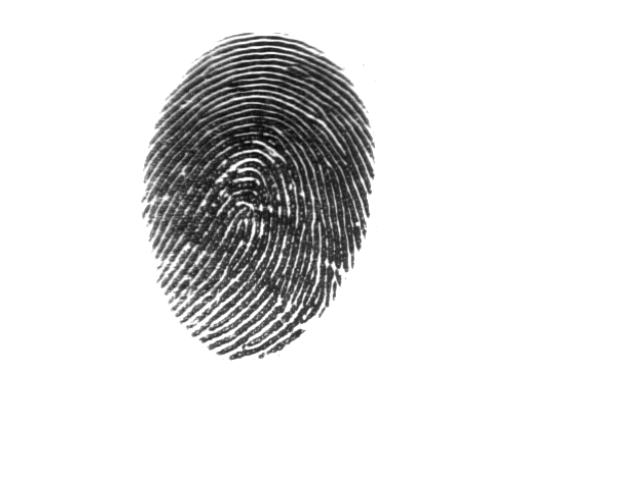} 
			\includegraphics[width=0.1\textwidth]{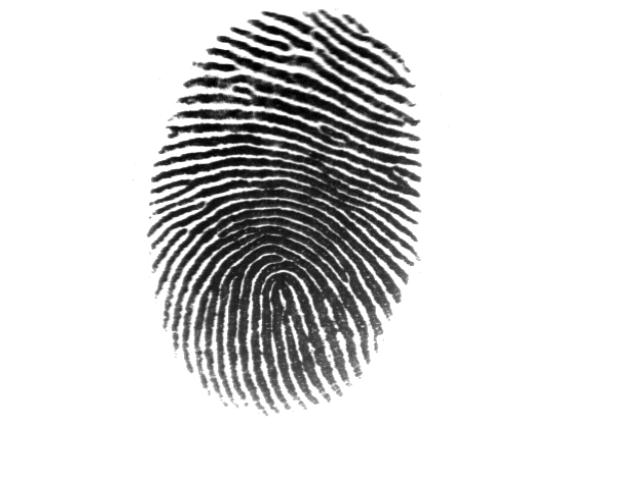} 
			\includegraphics[width=0.1\textwidth]{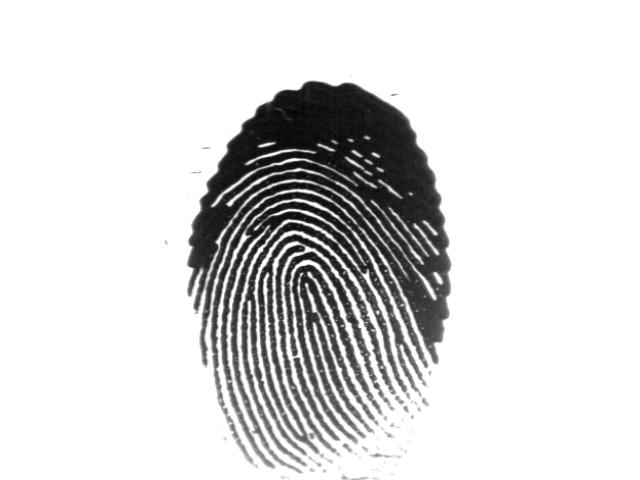} 
			\label{fig:second_sub}

		}			
		\subfigure[]
		{
			
			\includegraphics[width=0.1\textwidth]{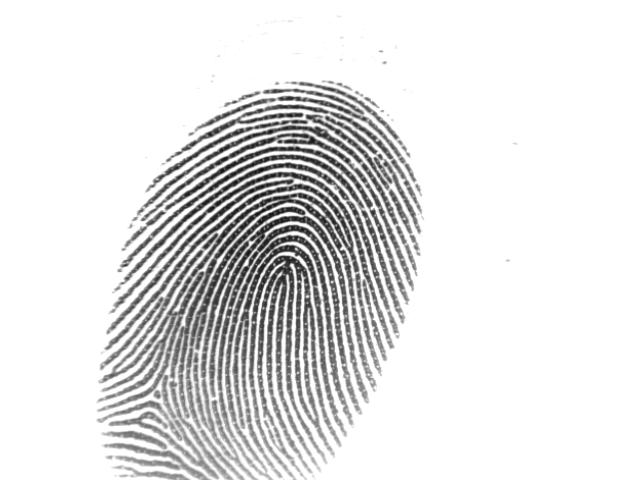} 
			\includegraphics[width=0.1\textwidth]{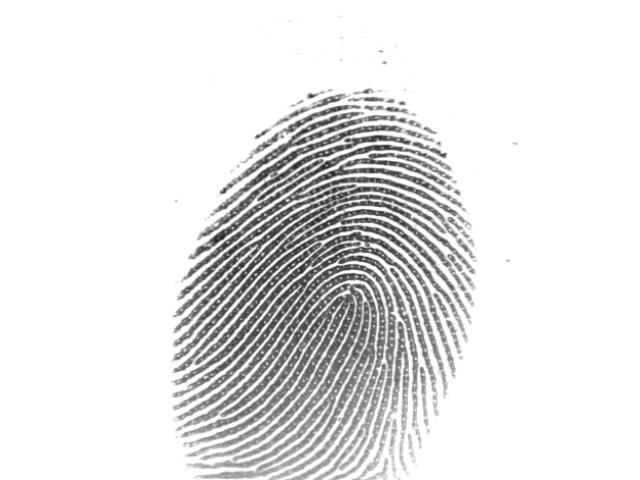} 
			\includegraphics[width=0.1\textwidth]{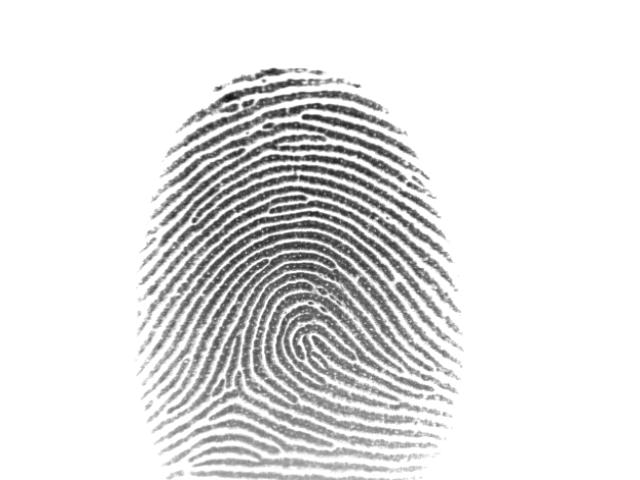} 
			\includegraphics[width=0.1\textwidth]{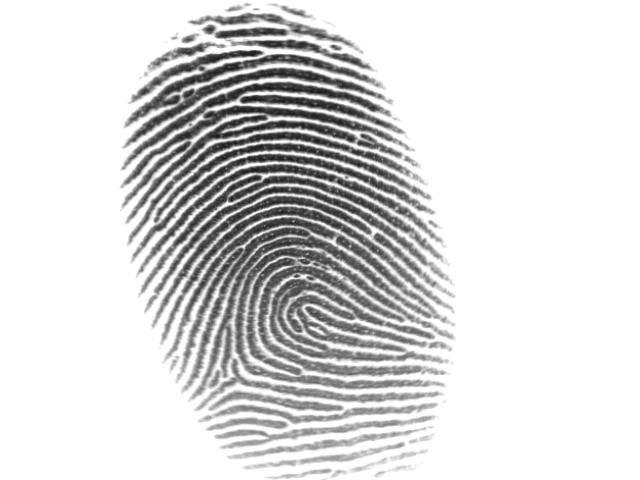} 
			\label{fig:second_sub}

		}

		\caption{Fingerprint images obtained in different clusters using fuzzy c-means: (a) Dry (b) Wet (c) Normal Dry (d) Normal Wet (e) Good }
		\label{clustimages}
	\end{figure}

First, we explore the experimental results for the first phase of FQA for the FVC2004 databases. All the images of each dataset $DB1-DB4$ are clustered manually in dry, normal dry, good, normal wet and wet quality classes. An objective method is proposed for performance evaluation of fingerprint quality assessment to test the efficacy of the FQA. Fingerprint images obtained after fuzzy c-means clustering in different classes $Q_c$ are compared with the manually clustered datasets in $Q_c$ classes. Figure \ref{clustimages} show the fingerprint images of different quality class obtained using fuzz c-means clustering.

\begin{table}[b]
	\centering
	\caption{Results of fingerprint quality clustering for FVC2004 $DB1$ database}
\label{DB1Aclustresult}	

  \resizebox{0.6\textwidth}{!}{
		\begin{tabular}{|c|c|c|c|c|c|c|c|}
			\hline
			& \multicolumn{7}{c|}{\textbf{Fuzzy C-Means Clustering}}                                                                                                                                                                                          \\ \hline
			\multirow{10}{*}{\textbf{\begin{tabular}[c]{@{}c@{}}Subjective\\ Clustering\end{tabular}}} &    \textbf{\begin{tabular}[c]{@{}c@{}}Quality \\ Clusters\end{tabular}}                            & \textit{Dry} & \textit{\begin{tabular}[c]{@{}c@{}}Normal \\ Dry\end{tabular}} & \textit{Good} & \textit{\begin{tabular}[c]{@{}c@{}}Normal \\ Wet\end{tabular}} & \textit{Wet} & \textit{Total}\\ \cline{2-8} \hline
			& \textit{Dry}                                                    & \textbf{139}          & 0                                                              & 0             & 0                                                              & 0       & 139     \\ \cline{2-8} 
			& \textit{\begin{tabular}[c]{@{}c@{}}Normal \\ Dry\end{tabular}}  & 2            & \textbf{192}                                                           & 4           & 0                                                              & 0  & 198         \\ \cline{2-8} 
			& \textit{Good}                                                   & 1            & 10                                                             & \textbf{135}           & 10                                                             & 0        & 156    \\ \cline{2-8} 
			& \textit{\begin{tabular}[c]{@{}c@{}}Normal \\ Wet\end{tabular}}  & 0            & 0                                                              & 7            & \textbf{151}                                                            & 0        & 158   \\ \cline{2-8} 
			& \textit{Wet}                                                    & 0            & 0                                                              & 1             & 1                                                              & \textbf{147}    & 149  \\ \cline{2-8}      
			& \textit{Total}                                                   & 142           & 202                                                            & 147           & 162                                                            & 147       & 800   \\ \hline
			& \textbf{\begin{tabular}[c]{@{}c@{}}Error\\  Rates\end{tabular}} & 2.11\%       & 5.05\%                                                        & 7.69\%       & 6.96\%                                                        & 0\%    &    4.50\%  \\ \hline
		\end{tabular}}	
	\end{table}

	\begin{table}[b]
		\centering
		\caption{Results of fingerprint quality clustering for FVC2004 $DB2$ database}
	\label{clustdb2_a}		

		\resizebox{0.6\textwidth}{!}{
			\begin{tabular}{|c|c|c|c|c|c|c|c|}
				\hline
				& \multicolumn{7}{c|}{\textbf{Fuzzy C-Means Clustering}}                                                                                                                                                                                          \\ \hline
				\multirow{10}{*}{\textbf{\begin{tabular}[c]{@{}c@{}}Subjective\\ Clustering\end{tabular}}} &   \textbf{\begin{tabular}[c]{@{}c@{}}Quality \\ Clusters\end{tabular}}                                                                 & \textit{Dry} & \textit{\begin{tabular}[c]{@{}c@{}}Normal \\ Dry\end{tabular}} & \textit{Good} & \textit{\begin{tabular}[c]{@{}c@{}}Normal \\ Wet\end{tabular}} & \textit{Wet} & \textit{Total}\\ \cline{2-8} \hline
				& \textit{Dry}                                                    & \textbf{211}          & 1                                                              & 0             & 0                                                              & 0       & 212     \\ \cline{2-8} 
				& \textit{\begin{tabular}[c]{@{}c@{}}Normal \\ Dry\end{tabular}}  & 1           & \textbf{207}                                                           & 3           & 1                                                              & 0  & 212         \\ \cline{2-8} 
				& \textit{Good}                                                   & 0            & 2                                                            & \textbf{175}           & 9                                                             & 0        & 186    \\ \cline{2-8} 
				& \textit{\begin{tabular}[c]{@{}c@{}}Normal \\ Wet\end{tabular}}  & 0            & 0                                                              & 6            & \textbf{125}                                                           & 2        & 133   \\ \cline{2-8} 
				& \textit{Wet}                                                    & 0            & 0                                                              & 1             & 2                                                              & \textbf{54}    & 57  \\ \cline{2-8}      
				& \textit{Total}                                                   & 212           & 210                                                            & 185           & 137                                                            & 56       & 800   \\ \hline
				& \textbf{\begin{tabular}[c]{@{}c@{}}Error\\  Rates\end{tabular}} & 0.47\%       & 2.35\%                                                        & 5.91\%       & 6.01\%                                                        & 5.26\%    &    3.50\%  \\ \hline
			\end{tabular}}	
		\end{table}
		
		\begin{table}[h]
			\centering
			\caption{Results of fingerprint quality clustering for FVC2004 $DB3$ database}
\label{clustdb3_a}			

			\resizebox{0.6\textwidth}{!}{
				\begin{tabular}{|c|c|c|c|c|c|c|c|}
					\hline
					& \multicolumn{7}{c|}{\textbf{Fuzzy C-Means Clustering}}                                                                                                                                                                                          \\ \hline
					\multirow{10}{*}{\textbf{\begin{tabular}[c]{@{}c@{}}Subjective\\ Clustering\end{tabular}}} &  \textbf{\begin{tabular}[c]{@{}c@{}}Quality \\ Clusters\end{tabular}}                                                                  & \textit{Dry} & \textit{\begin{tabular}[c]{@{}c@{}}Normal \\ Dry\end{tabular}} & \textit{Good} & \textit{\begin{tabular}[c]{@{}c@{}}Normal \\ Wet\end{tabular}} & \textit{Wet} & \textit{Total}\\ \cline{2-8} \hline
					& \textit{Dry}                                                    & \textbf{61}          & 2                                                              & 1             & 0                                                              & 0       &64     \\ \cline{2-8} 
					& \textit{\begin{tabular}[c]{@{}c@{}}Normal \\ Dry\end{tabular}}  & 2            & \textbf{186}                                                           & 9           & 0                                                              & 0  & 197        \\ \cline{2-8} 
					& \textit{Good}                                                   & 1            & 10                                                             & \textbf{254}          & 5                                                             & 0        & 270    \\ \cline{2-8} 
					& \textit{\begin{tabular}[c]{@{}c@{}}Normal \\ Wet\end{tabular}}  & 0            & 0                                                              & 2            & \textbf{73}                                                            & 3        & 78   \\ \cline{2-8} 
					& \textit{Wet}                                                    & 0            & 0                                                              & 1             & 1                                                              & \textbf{189}    & 191  \\ \cline{2-8}      
					& \textit{Total}                                                   & 64           & 198                                                            & 267           & 79                                                            & 192       & 800   \\ \hline
					& \textbf{\begin{tabular}[c]{@{}c@{}}Error\\  Rates\end{tabular}} & 4.68\%       & 6.06\%                                                        & 4.86\%       & 7.59\%                                                        & 1.56\%    &    4.62\%  \\ \hline
				\end{tabular}}
	
			\end{table}

			\begin{table}[h]
				\centering
				\caption{Results of fingerprint quality clustering for FVC2004 $DB4$ database}	
	\label{clustdb4_a}				
	
				\resizebox{0.6\textwidth}{!}{
					\begin{tabular}{|c|c|c|c|c|c|c|c|}
						\hline
						& \multicolumn{7}{c|}{\textbf{Fuzzy C-Means Clustering}}                                                                                                                                                                                          \\ \hline
						\multirow{10}{*}{\textbf{\begin{tabular}[c]{@{}c@{}}Subjective\\ Clustering\end{tabular}}} &  \textbf{\begin{tabular}[c]{@{}c@{}}Quality \\ Clusters\end{tabular}}     & \textit{Dry} & \textit{\begin{tabular}[c]{@{}c@{}}Normal \\ Dry\end{tabular}} & \textit{Good} & \textit{\begin{tabular}[c]{@{}c@{}}Normal \\ Wet\end{tabular}} & \textit{Wet} & \textit{Total}\\ \cline{2-8} \hline
						& \textit{Dry}                                                    & \textbf{151}          & 0                                                              & 0             & 0                                                              & 0       & 151     \\ \cline{2-8} 
						& \textit{\begin{tabular}[c]{@{}c@{}}Normal \\ Dry\end{tabular}}  & 4           & \textbf{227}                                                           & 0           & 0                                                              & 0  & 231         \\ \cline{2-8} 
						& \textit{Good}                                                   & 0            & 1                                                            & \textbf{209}           & 1                                                             & 1        & 212    \\ \cline{2-8} 
						& \textit{\begin{tabular}[c]{@{}c@{}}Normal \\ Wet\end{tabular}}  & 0            & 0                                                              & 22            & \textbf{142}                                                           & 4        & 168   \\ \cline{2-8} 
						& \textit{Wet}                                                    & 0            & 0                                                              & 0             & 0                                                             & \textbf{38}    & 38  \\ \cline{2-8}      
						& \textit{Total}                                                   & 155           & 228                                                           & 231           & 143                                                            & 43       & 800   \\ \hline
						& \textbf{\begin{tabular}[c]{@{}c@{}}Error\\  Rates\end{tabular}} &2.58\%       & 0.43\%                                                        & 9.52\%       & 0.69\%                                                        & 11.62\%    &   4.12\%  \\ \hline
					\end{tabular}}

				\end{table}

				The clustering error in each quality class $Q_c$ for FVC2004 $DB1$ dataset is reported in Table \ref{DB1Aclustresult}. Error rates for individual quality class is determined as number of wrong classified fingerprint images divided by total images in that quality cluster. Fuzzy c-means clusters 142 fingerprint images in the dry quality class, out of which 139 are predicted correctly and 3 (2-normal dry, 1-good) are predicted in wrong quality class. Hence, error in clustering of dry images is $3/142\times100 =2.11\%$. Similarly, number of wrong clustered images for normal dry, good, normal wet and wet class is 10 (good), 12 (4-normal dry, 7-normal wet, 1-wet), 11(10-good, 1-wet) and 0, which causes error rates of 5.05\%, 7.69\%, 6.96\% and 0\% respectively. The overall error rate is 4.50\% for the $DB1$ dataset of FVC2004. Experimental results exhibit that error in prediction of the correct quality class for dry and wet fingerprint images is very less as well as none of the dry images are classified as wet or vice-versa. On the contrary, error rates for normal dry and normal wet images are little high as good quality images contain both the normal dry and normal wet regions. It is observed that the images which are not clustered correctly are assigned to quality classes which are nearer to original quality class. As an instance of this observation, it can be seen in Table \ref{DB1Aclustresult} that the some of the images which are originally in the wet quality class are clustered wrongly in the normal wet (1) and good (1) quality class. None of the wet quality images (according to subjective clustering) is clustered in normal dry or dry class. Similar observations can be seen for other quality clusters of fingerprint images. 

				%
				%
				%

				Similar experiments are performed on $DB2-DB4$ datasets of FVC2004 to cluster the fingerprint images of these datasets in appropriate quality class. Results for these datasets are shown in Table \ref{clustdb2_a}, \ref{clustdb3_a} and \ref{clustdb4_a} respectively. Prediction of fingerprint images in the different quality classes for $DB2$ is done with error rates of 0.47\%, 2.35\%, 5.91\%, 6.01\% and 5.26\% for dry, normal dry, good, normal wet and wet clusters respectively (see Table \ref{clustdb2_a}). Overall error rate is 3.50\% for FVC2004 $DB2$ dataset. FVC2004 $DB3$ dataset is clustered in dry, normal dry, good, normal wet and wet quality classes with error rates of 4.68\%, 6.06\%, 4.86\%, 7.59\% and 1.56\% respectively which constitutes overall error rate of 4.62\% (see Table \ref{clustdb3_a}). Error rates for clustering FVC2004 $DB4$ dataset in appropriate quality clusters is 2.58\%, 0.43\%, 9.52\%, 0.69\% and 11.62\% for dry, normal dry, good, normal wet and wet quality classes respectively. Overall error rate is 4.12\% for FVC2004 $DB4$ dataset.
				
				\subsection{Experimental results of two-stage fingerprint quality enhancement}
				The proposed fingerprint enhancement scheme comprises of two consecutive stages of enhancement as shown in Figure \ref{blockdiagram}. In the first stage enhancement, fingerprint images undergoes through QAP to clear the ridge-valley pattern of fingerprints in low quality (dry, normal dry, normal wet and wet) images. Further, fingerprint images are enhanced with some well known enhancement algorithms in second stage of enhancement. Experimental results of the two stage enhancement are reported in the following section.

		\begin{figure}[t]
			\centering
			
			\subfigure[]
			{
				\includegraphics[width=0.7in]{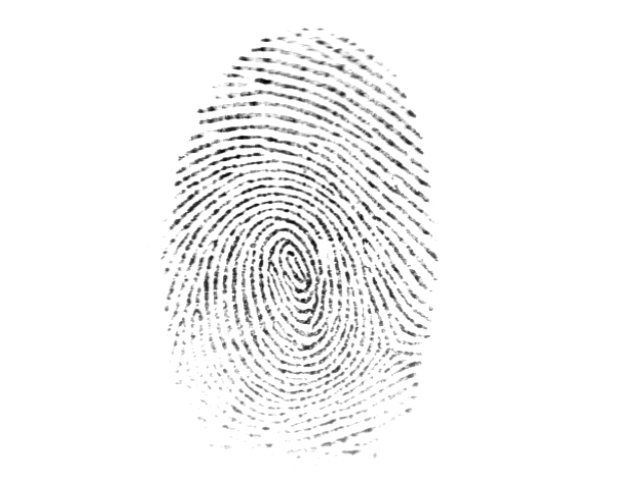}
				\includegraphics[width=0.7in]{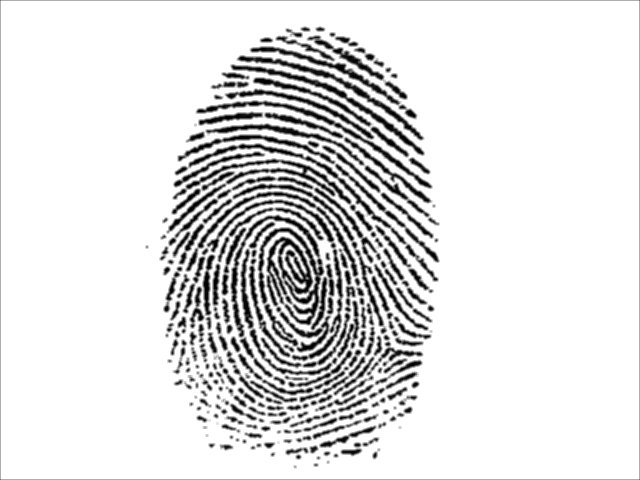}
				\label{Qap dry}
			}
			\hspace{0.02mm}
			\subfigure[]
			{
				\includegraphics[width=0.7in]{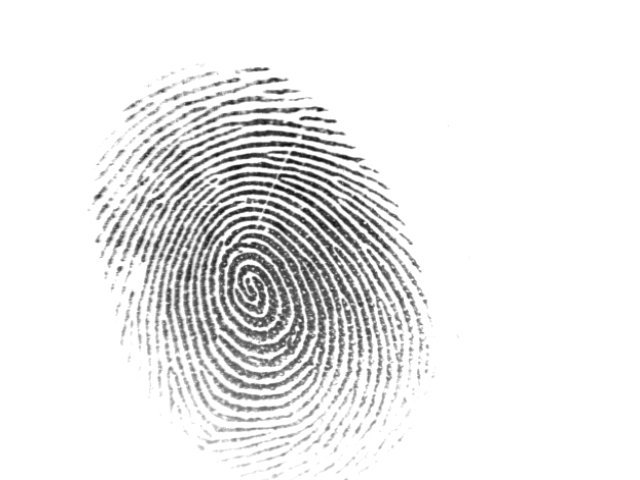}
				\includegraphics[width=0.7in]{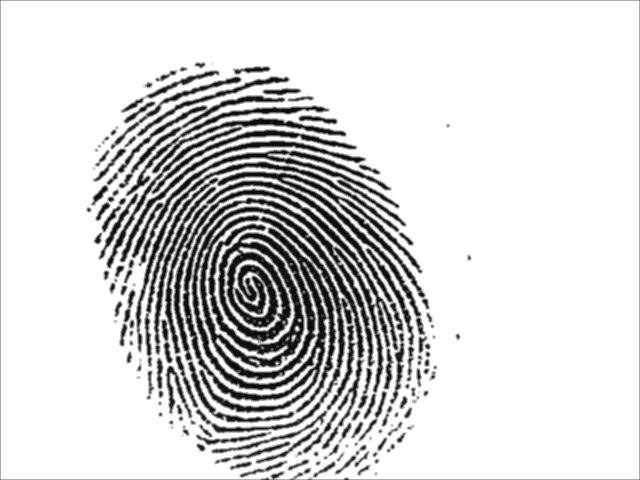}
				\label{Qap nd}
			}		
			\subfigure[]
			{
				\includegraphics[width=0.7in]{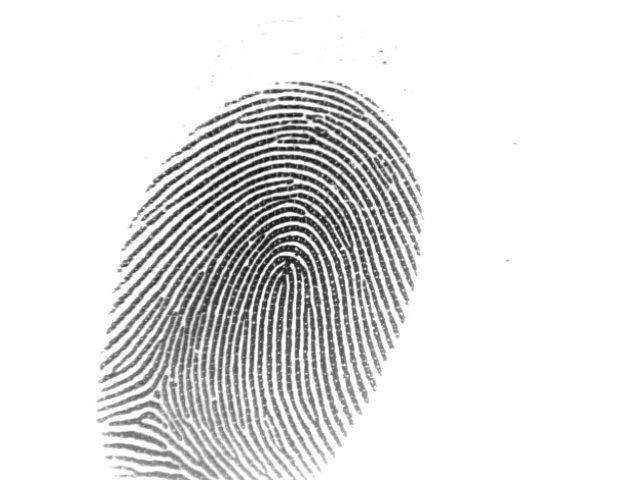}
				\includegraphics[width=0.7in]{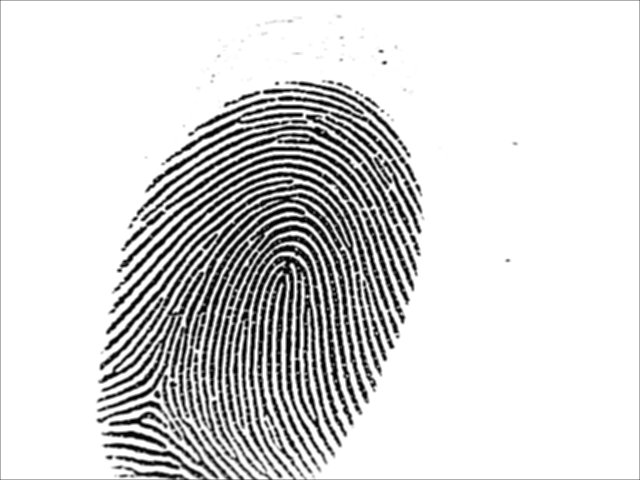}
				\label{Qap good}
			}		
			\hspace{0.02mm}
			\subfigure[]
			{
				\includegraphics[width=0.7in]{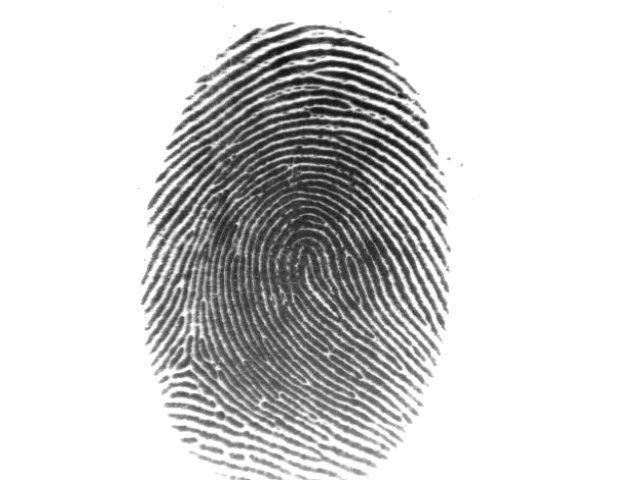}
				\includegraphics[width=0.7in]{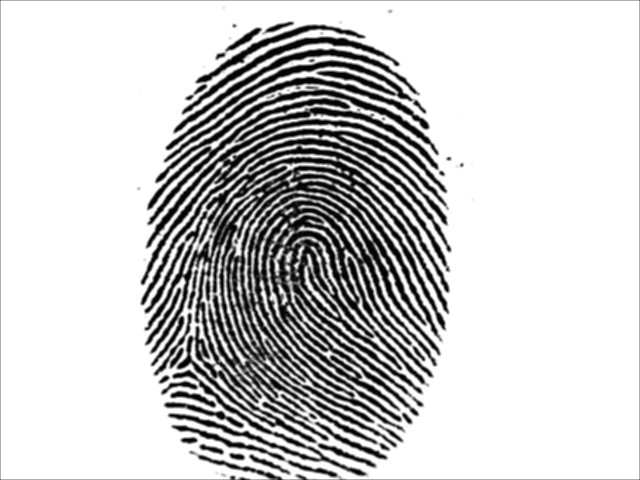}
				\label{Qap nw}
			}		
			\hspace{2mm}
			\subfigure[]
			{
				\includegraphics[width=0.7in]{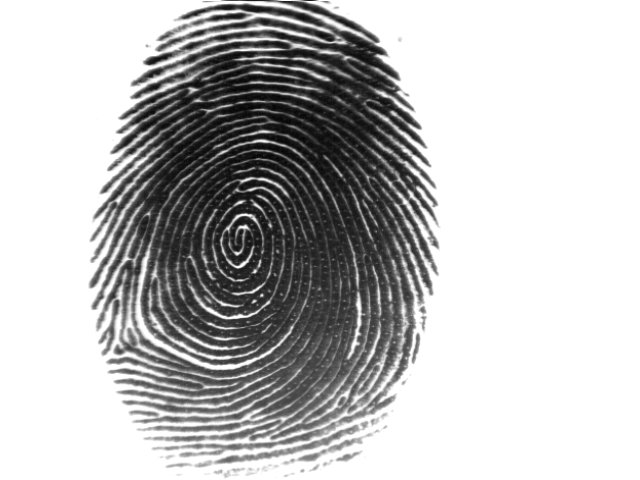}
				\includegraphics[width=0.7in]{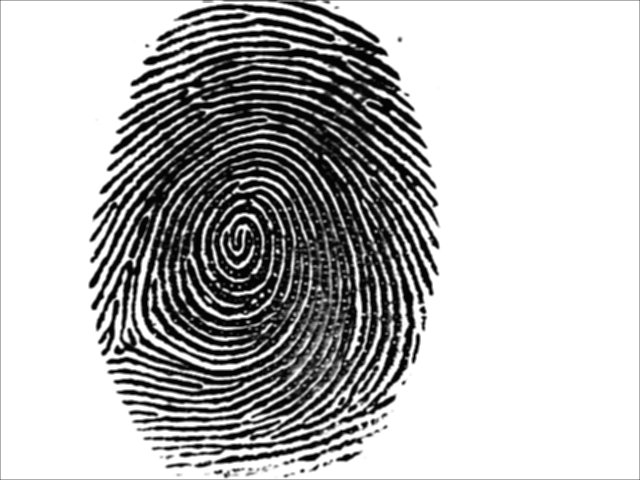}
				\label{Qap wet}
			}

			\caption{Quality adaptive preprocessing of images from FVC2004 $DB1$ database: (a) Dry (b) Normal dry (c) Good (d) Normal wet (e) Wet ; the columns correspond to 1) the original image 2) preprocessed image}
			\label{qap}
		\end{figure}							
				\subsubsection{Quality adaptive preprocessing}\label{QAP}
				After fingerprint quality assessment using fuzzy c-means in the phase 1 of the proposed method, a quality adaptive preprocessing (QAP) of fingerprint images of each quality cluster $Q_c$ is performed. Original and enhanced images using QAP from each quality class are shown in Figure \ref{qap}. To assess the efficiency of the proposed quality adaptive preprocessing method, seven-dimensional feature vector $F_s$ is extracted from subjectively clustered 200 random fingerprint images of all the five clusters ($Q_c$) ($5\times200=1000$ fingerprint images) from $DB1-DB4$. The feature vector of these 1000 fingerprint images is fed as input to train a decision tree model for predicting the fingerprint quality class. Decision Tree (DT) is a non-parametric supervised learning method used for classification problems. Its goal is to create a classification model that predicts the target variable or class by learning simple decision rules inferred from the data features. DT has an advantage that they can handle multi-class classification problems and it is easy to visualize and understand the classification rules. Further, another 50 fingerprint images (different from 1000 fingerprint images used for training) from each of the quality clusters are selected to show the improvement in the fingerprint quality. Quality class of these 50 fingerprint images of each class are predicted using trained decision tree model. Table \ref{beforeqap} and \ref{afterqap} represent the fingerprint quality class prediction before and after the QAP respectively.
	\vspace{-2mm}
\begin{table}[H]
	\centering
			\caption{Fingerprint quality class prediction before QAP}
\label{beforeqap}	

	\resizebox{0.65\textwidth}{!}{
		\begin{tabular}{|c|c|c|c|c|c|c|}
			\hline
			& \multicolumn{6}{c|}{\textbf{Decision Tree Classification}}                                                            \\ \hline
			&   \textbf{\begin{tabular}[c]{@{}c@{}}Quality \\ Clusters\end{tabular}}            & Dry         & Normal Dry  & Good        & Normal Wet  & Wet         \\ \hline
			\multirow{5}{*}{\textbf{\begin{tabular}[c]{@{}c@{}}Subjective \\ Quality\end{tabular}}} & Dry        & \textbf{46} & 4           & 0           & 0           & 0           \\ \cline{2-7} 
			& Normal Dry & 0           & \textbf{50} & 0           & 0           & 0           \\ \cline{2-7} 
			& Good       & 0           & 2           & \textbf{48} & 0           & 0           \\ \cline{2-7} 
			& Normal Wet & 0           & 0           & 3           & \textbf{46} & 1           \\ \cline{2-7} 
			& Wet        & 0           & 0           & 0           & 2           & \textbf{48} \\ \hline
		\end{tabular}}
	\vspace{-2mm}
	\end{table}
	\vspace{-2mm}	
	
	\begin{table}[H]
		\centering
			\caption{Fingerprint quality class prediction after QAP}
\label{afterqap}		
	
		\resizebox{0.65\textwidth}{!}{
			\begin{tabular}{|c|c|c|c|c|c|c|}
				\hline
				& \multicolumn{6}{c|}{\textbf{Decision Tree Classification}}                                                            \\ \hline
				&  \textbf{\begin{tabular}[c]{@{}c@{}}Quality \\ Clusters\end{tabular}}           & Dry         & Normal Dry  & Good        & Normal Wet  & Wet         \\ \hline
				\multirow{5}{*}{\textbf{\begin{tabular}[c]{@{}c@{}}Subjective \\ Quality\end{tabular}}} & Dry        & \textbf{3} & 47           & 0           & 0           & 0           \\ \cline{2-7} 
				& Normal Dry & 0           & \textbf{22} & 27          & 1           & 0           \\ \cline{2-7} 
				& Good       & 0           & 2           & \textbf{48} & 0           & 0           \\ \cline{2-7} 
				& Normal Wet & 0           & 17           &21          & \textbf{12} &            \\ \cline{2-7} 
				& Wet        & 0           & 0           & 2           & 48           & \textbf{0} \\ \hline
			\end{tabular}}
		\end{table}	
		
				Results of the DT quality prediction before QAP in Table \ref{beforeqap} show that there are 46 dry and 4 normal dry fingerprint images out of the 50 manually selected dry fingerprint images. After processing the fingerprint images with QAP, DT predicts 47 dry images in normal dry class while 3 remains in dry class in Table \ref{afterqap}. This shows that after QAP, quality of most of the fingerprint images of the dry class is improved and they are moved in normal dry class. A similar improvement is seen for wet fingerprint images in Table \ref{beforeqap}, which are advanced to normal wet (48) and good (2) quality class after QAP in Table \ref{afterqap}. The quality of normal dry and normal wet fingerprint images are also improved significantly as some fingerprint images from these classes are moved in good quality class after preprocessing them with QAP. None of the good quality images are moved to any low class as good quality images are preprocessed with optimal parameter values in QAP. These results exhibits significant improvement in the  fingerprint images of each quality class while the quality of the good fingerprint images remains the same. 
\begin{figure*}[]
	\centering
	\resizebox{\textwidth}{!}{
	\subfigure[]
	{
		\includegraphics[width=0.8in]{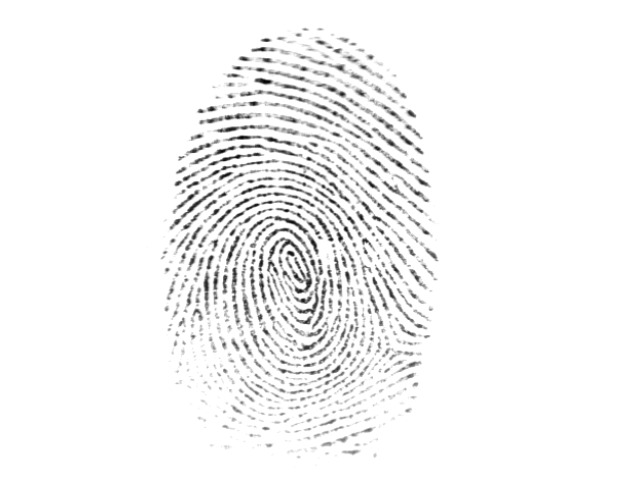}
		\includegraphics[width=0.8in]{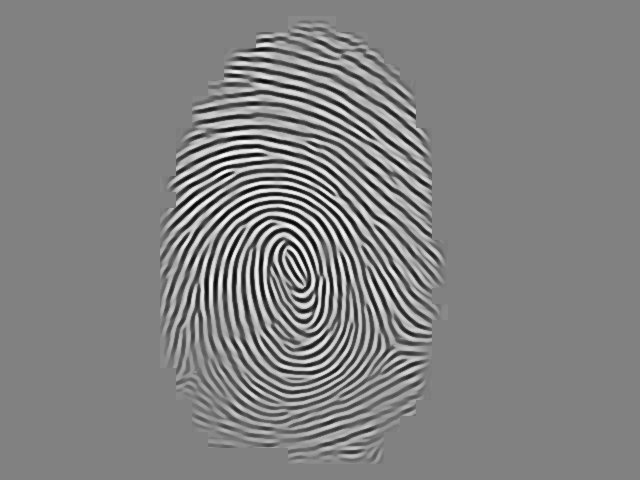}
		\includegraphics[width=0.8in]{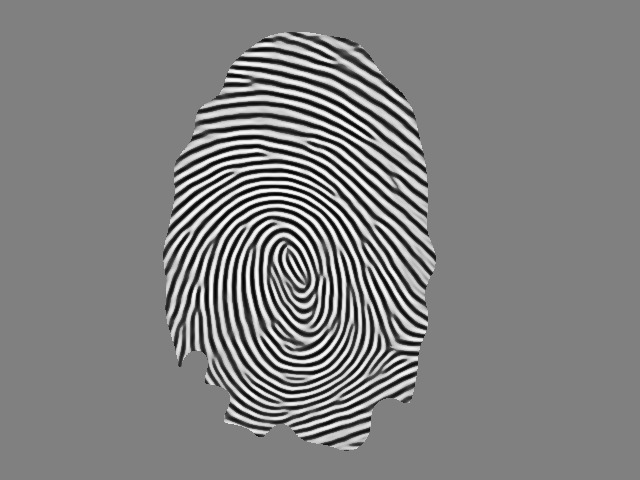}
		\includegraphics[width=0.8in]{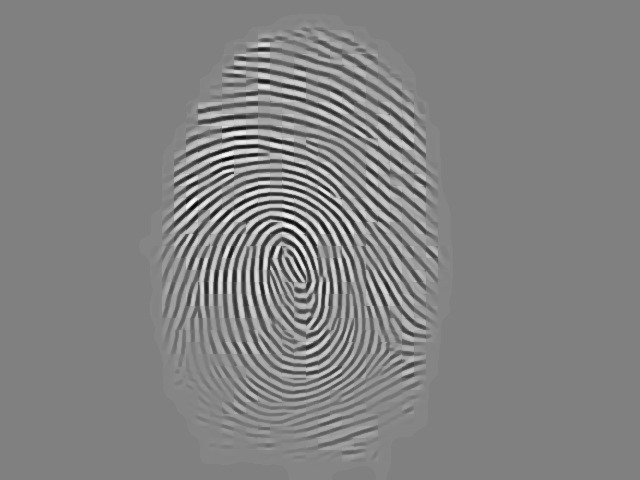}
		\includegraphics[width=0.8in]{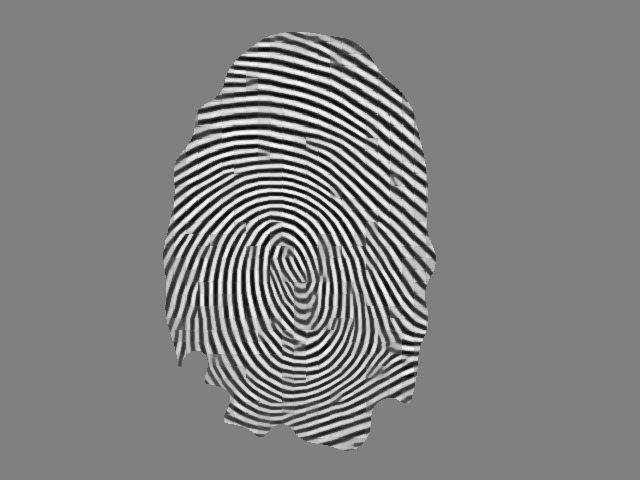}
		\includegraphics[width=0.8in]{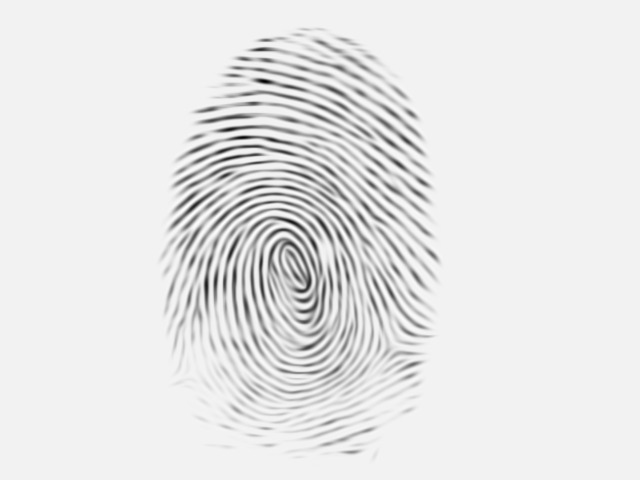}
		\includegraphics[width=0.8in]{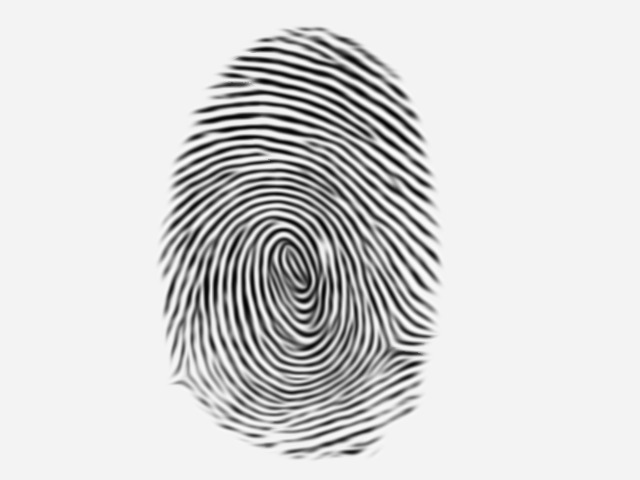}
		
	}}
	\resizebox{\textwidth}{!}{
	\subfigure[]
	{
		\includegraphics[width=0.8in]{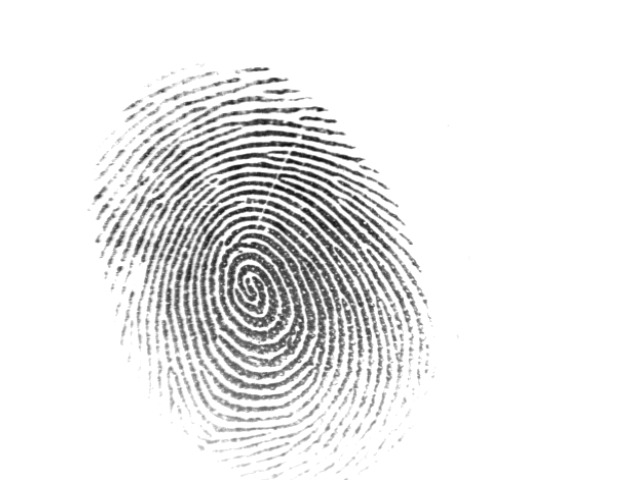}
		\includegraphics[width=0.8in]{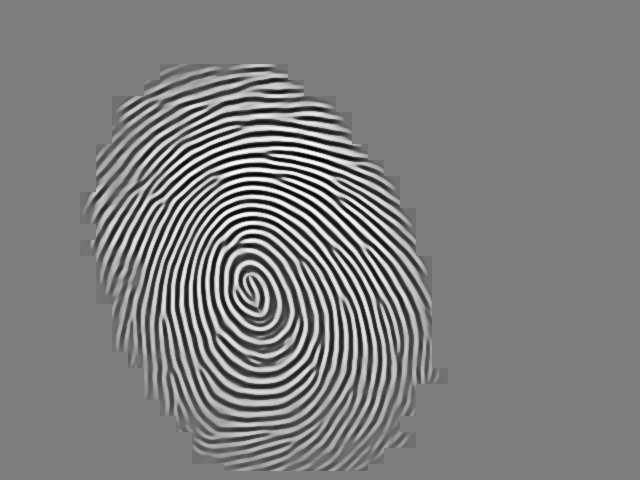}
		\includegraphics[width=0.8in]{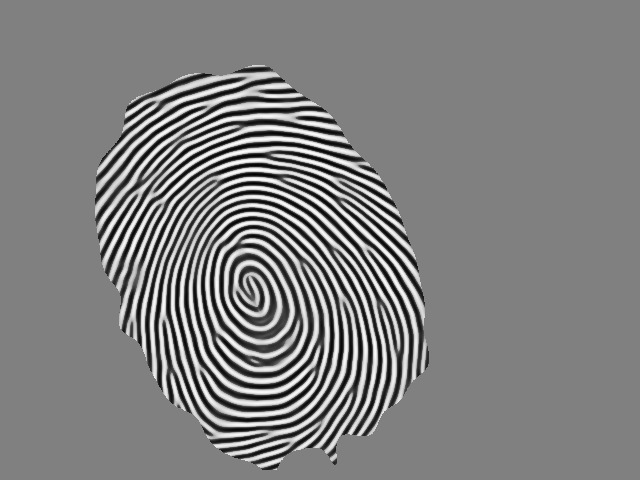}
		\includegraphics[width=0.8in]{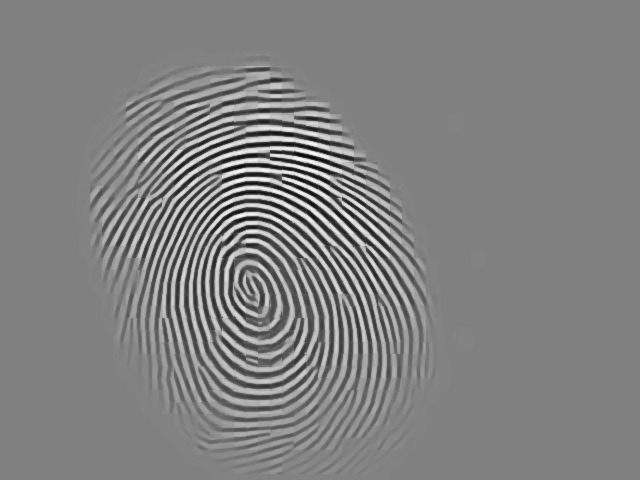}
		\includegraphics[width=0.8in]{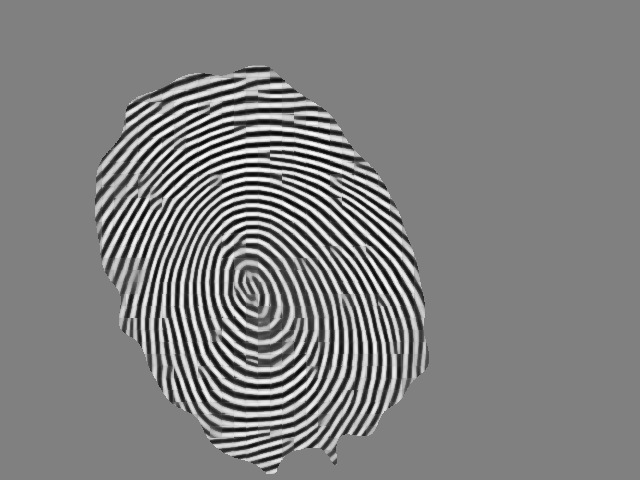}
		\includegraphics[width=0.8in]{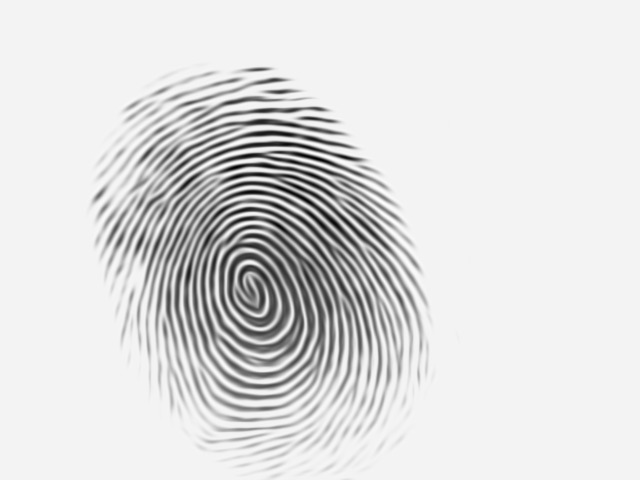}
		\includegraphics[width=0.8in]{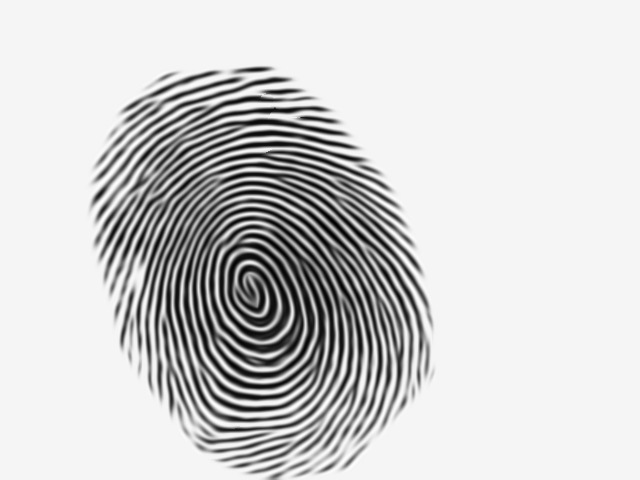}
	}}
	\resizebox{\textwidth}{!}{	
	\subfigure[]
	{	
		\includegraphics[width=0.8in]{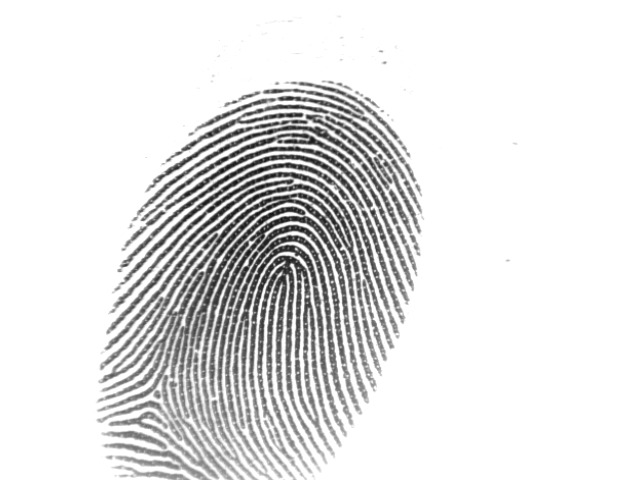}
		\includegraphics[width=0.8in]{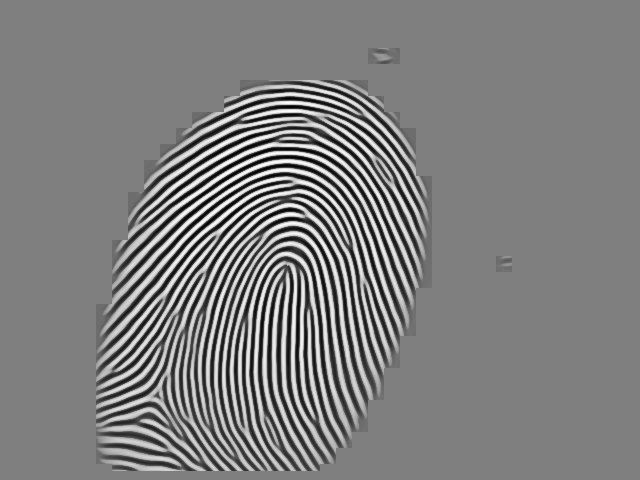}
		\includegraphics[width=0.8in]{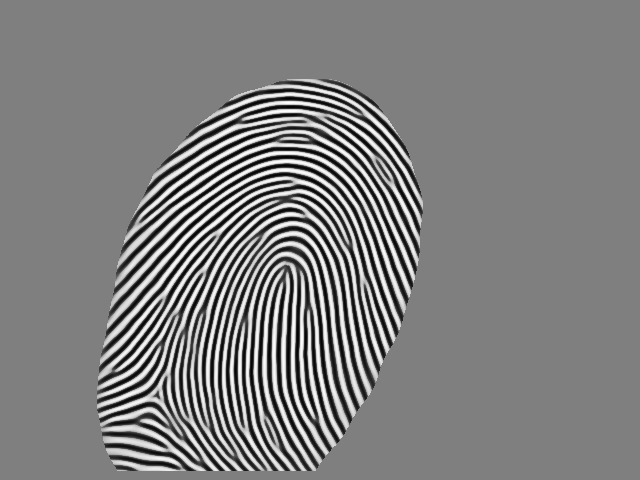}
		\includegraphics[width=0.8in]{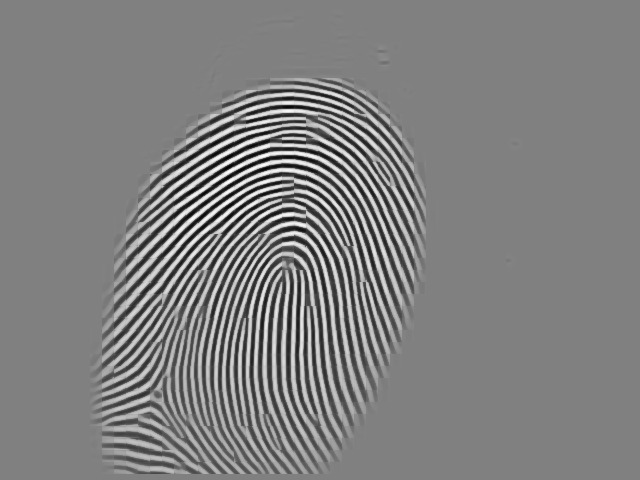}
		\includegraphics[width=0.8in]{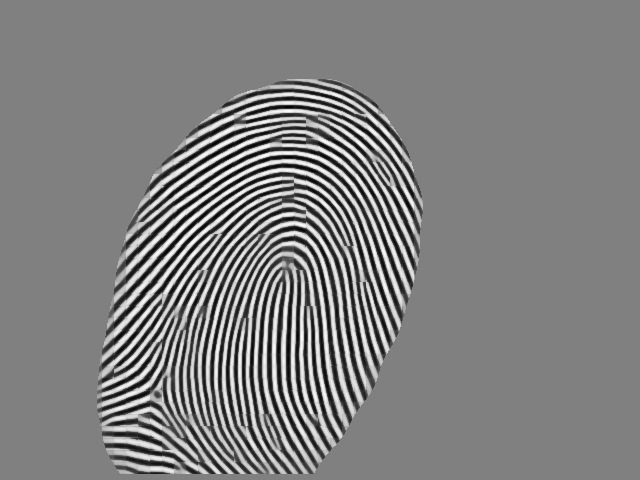}
		\includegraphics[width=0.8in]{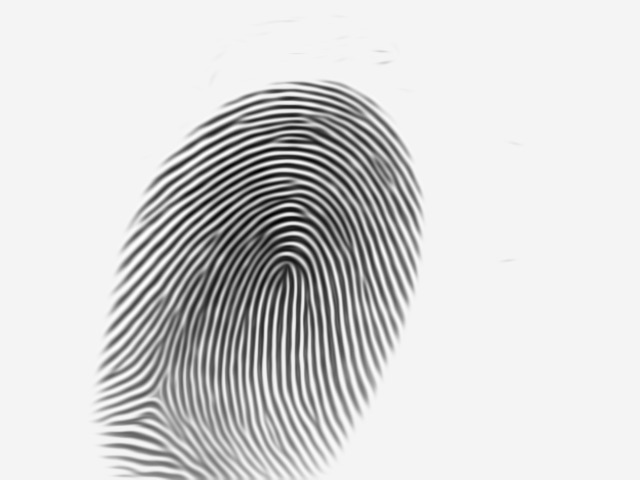}
		\includegraphics[width=0.8in]{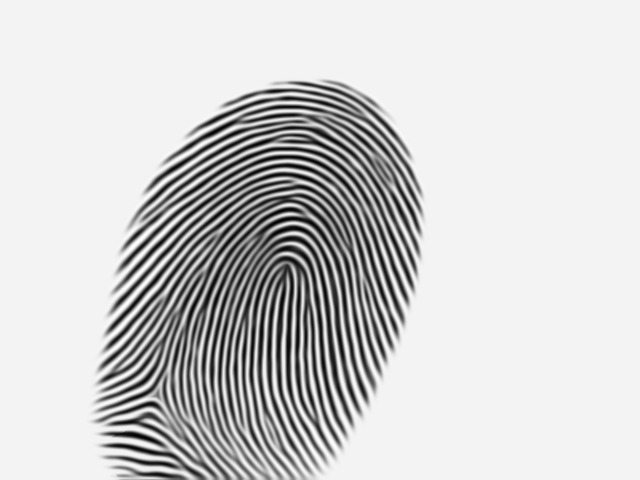}
	}}
	\resizebox{\textwidth}{!}{
	\subfigure[]
	{
		\includegraphics[width=0.8in]{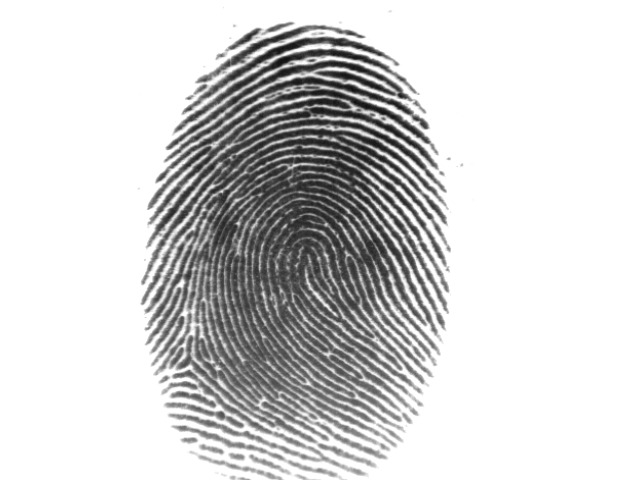}
		\includegraphics[width=0.8in]{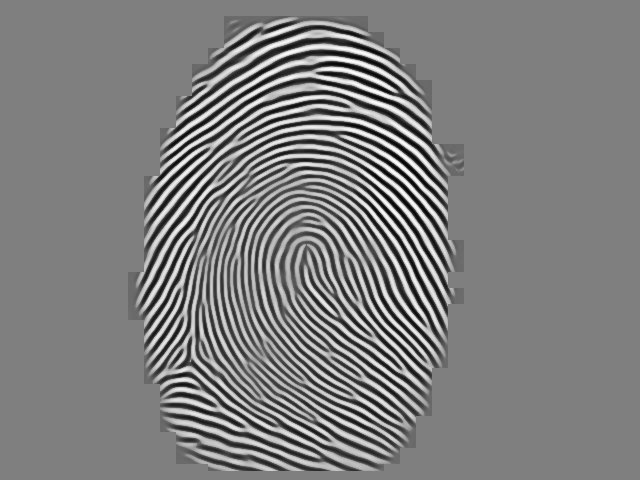}
		\includegraphics[width=0.8in]{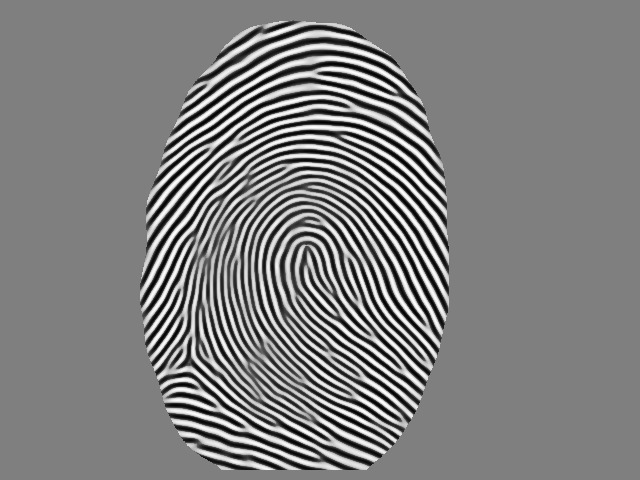}
		\includegraphics[width=0.8in]{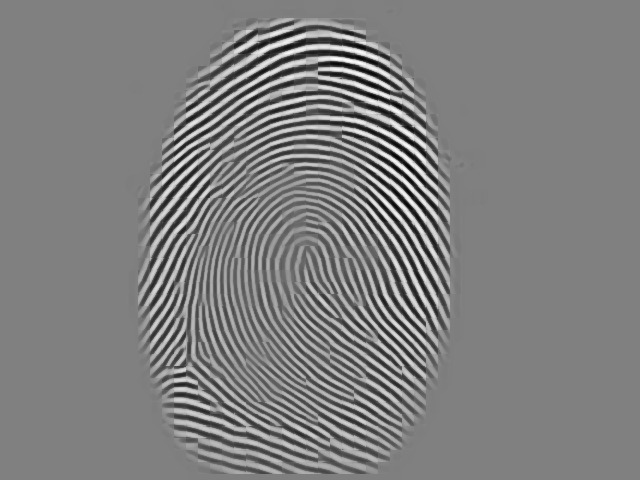}
		\includegraphics[width=0.8in]{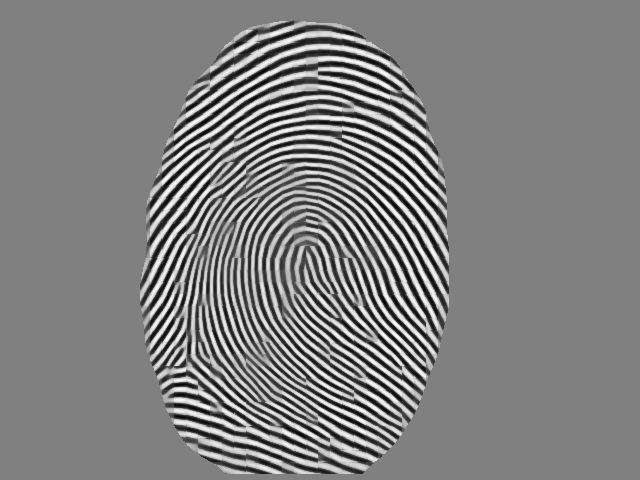}
		\includegraphics[width=0.8in]{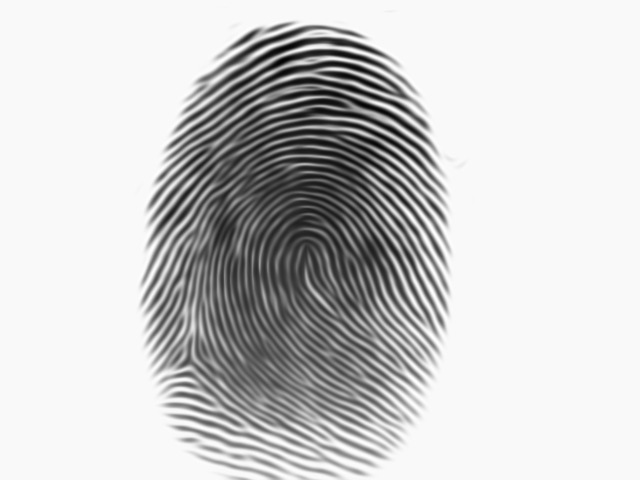}
		\includegraphics[width=0.8in]{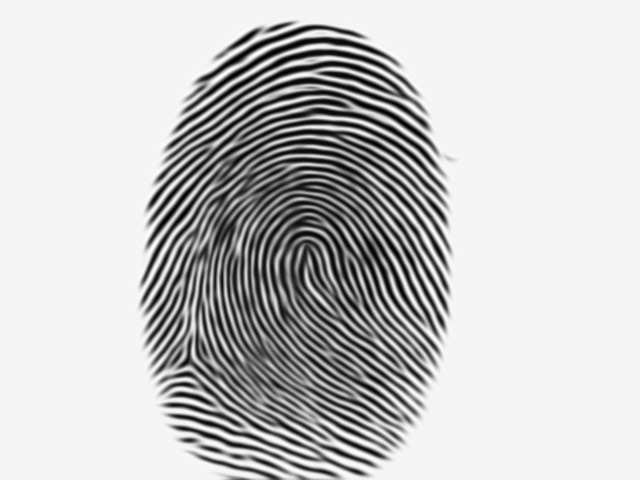}
	}}
	\resizebox{\textwidth}{!}{
	\subfigure[]
	{
		\includegraphics[width=0.8in]{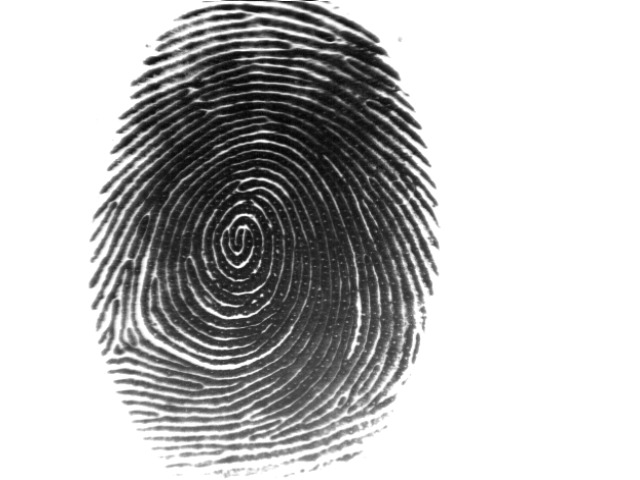}
		\includegraphics[width=0.8in]{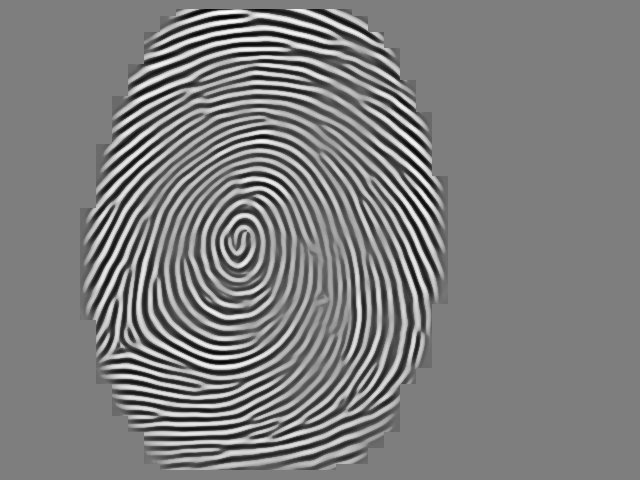}
		\includegraphics[width=0.8in]{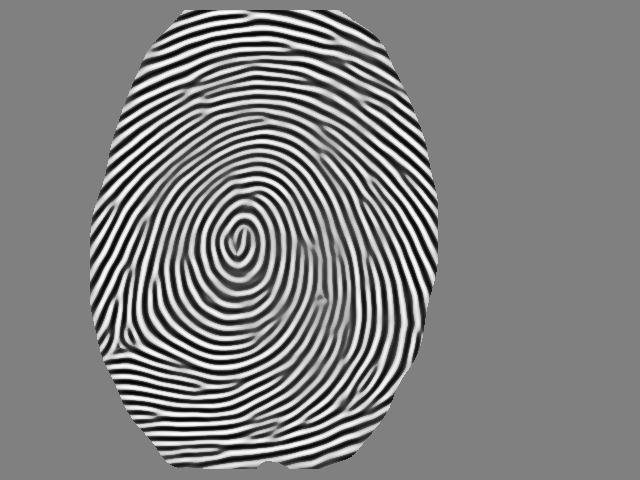}
		\includegraphics[width=0.8in]{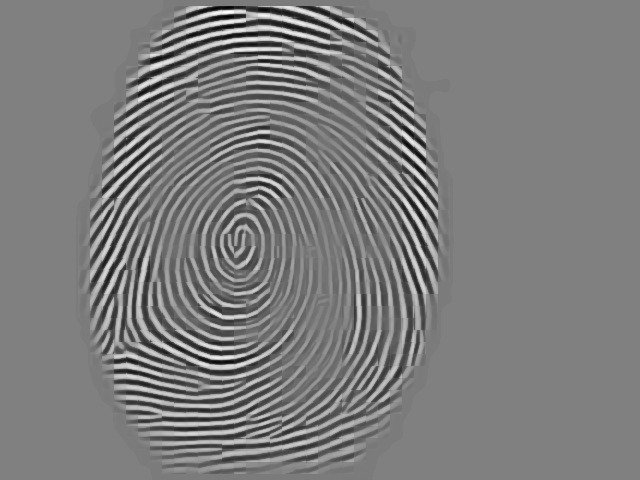}
		\includegraphics[width=0.8in]{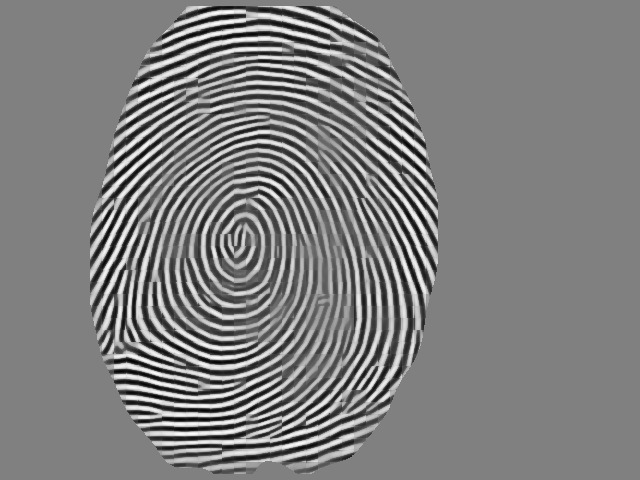}
		\includegraphics[width=0.8in]{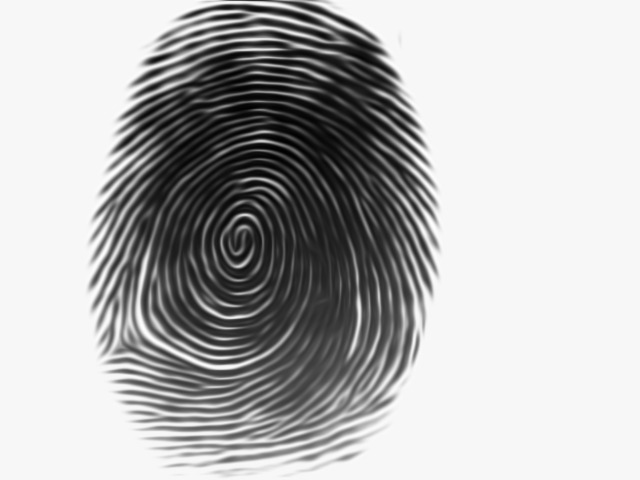}
		\includegraphics[width=0.8in]{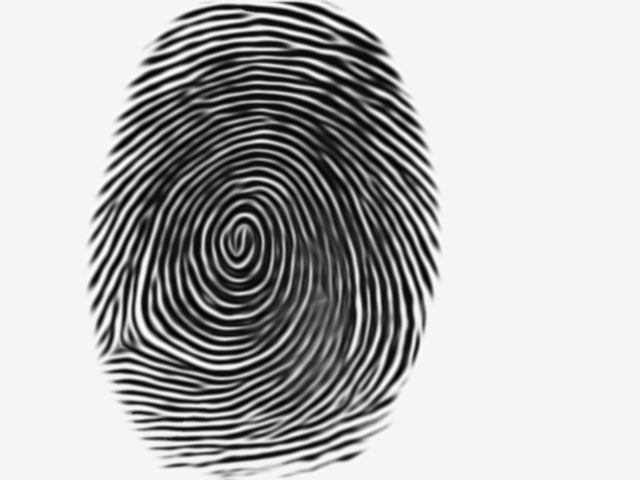}
}		}			
	\caption{Enhancement results for different quality fingerprint images from FVC2004 $DB1$ database. (a) Dry (b) Normal dry (c) Good (d) Normal wet (e) Wet ; the columns
		correspond to 1) the original image 2) Gabor enhancement 3) QAP + Gabor enhancement 4) STFT enhancement 5) QAP + STFT enhancement 6) ODF enhancement 7) QAP + ODF enhancement.}
	\label{enhanced}
\end{figure*}	
							
\begin{table}[h]
	\centering
		\caption{EER's obtained using NIST FIS2 matcher (MINDTCT and BOZORTH3) for QAP and without QAP based enhancement methods.}
	\label{eer}
	\resizebox{0.9\textwidth}{!}{
		\begin{tabular}{|c|c|c|c|c|c|c|c|}
			\hline
			\multicolumn{2}{|c|}{\textbf{Database}}          & \begin{tabular}[c]{@{}c@{}}No \\ pre-enhancement\end{tabular} & QAP              & \begin{tabular}[c]{@{}c@{}}Relative\\ improvement\end{tabular} & Gabor\cite{journals/pami/HongWJ98}    & QAP + Gabor      & \begin{tabular}[c]{@{}c@{}}Relative\\ improvement\end{tabular} \\ \hline
			\multirow{4}{*}{FVC 2004} & DB1 & 14.50 \%                                                      & 10.74 \%         & \textbf{25.93 \%}                                              & 16.90 \% & \textbf{9.12 \%} & \textbf{46.03 \%}                                              \\ \cline{2-8} 
			& DB2 & 9.50 \%                                                       & 9.31 \%          & \textbf{2.00 \%}                                               & 14.40 \% & 8.41 \%          & \textbf{41.59 \%}                                              \\ \cline{2-8} 
			& DB3 & 6.20 \%                                                       & 7.53 \%          & -21.45 \%                                                      & 7.10 \%  & 5.65 \%          & \textbf{20.42 \%}                                              \\ \cline{2-8} 
			& DB4 & \textbf{7.30 \%}                                              & 9.95 \%          & -36.30 \%                                                      & 9.80 \%  & 9.32 \%          & \textbf{4.89 \%}                                               \\ \hline
			&     & STFT\cite{Chikkerur2007198}                                                         & QAP + STFT       & \begin{tabular}[c]{@{}c@{}}Relative\\ improvement\end{tabular} & ODF\cite{Gottschlich2012}      & QAP + ODF        & \begin{tabular}[c]{@{}c@{}}Relative\\ improvement\end{tabular} \\ \hline
			\multirow{4}{*}{FVC 2004} & DB1 & 19.10 \%                                                      & 9.43 \%          & \textbf{50.62 \%}                                              & 9.55 \%  & 8.55 \%          & \textbf{10.47 \%}                                              \\ \cline{2-8} 
			& DB2 & 11.90 \%                                                      & \textbf{8.27 \%} & \textbf{30.50 \%}                                              & 9.06 \%  & 8.92 \%          & \textbf{1.54 \%}                                               \\ \cline{2-8} 
			& DB3 & 7.60 \%                                                       & \textbf{5.40 \%} & \textbf{28.94 \%}                                              & 6.69 \%  & 6.16 \%          & \textbf{7.92 \%}                                               \\ \cline{2-8} 
			& DB4 & 10.90 \%                                                      & 9.55 \%          & \textbf{12.38 \%}                                              & 9.98 \%  & 9.57 \%          & \textbf{4.10 \%}                                               \\ \hline
	\end{tabular}}

\end{table}

\subsubsection{Gabor/STFT/ODF enhancement}
In order to show that the proposed first stage QAP process benefit the second stage enhancement using Gabor/STFT/ODF enhancement algorithms, experiments with QAP as front-end for them are conducted. Therefore, QAP processed fingerprint images of $DB1-DB4$ datasets are further enhanced with Gabor enhancement \cite{journals/pami/HongWJ98} using morphological segmentation \cite{Fahmy2013}, STFT enhancement \cite{Chikkerur2007198} and ODF based enhancement \cite{Gottschlich2012} method. Enhanced fingerprint images of different quality  are shown in Figure \ref{enhanced} by using the combination of QAP + Gabor/STFT/ODF enhancements with traditional single stage Gabor/STFT/ODF enhancements. We can see from the Figure \ref{enhanced} that fingerprint images enhanced with QAP based methods (columns 3, 5 and 7 of Figure \ref{enhanced}) shows better visual results with respect to traditional Gabor/STFT/ODF enhancements (columns 2, 4 and 6 of Figure \ref{enhanced}).  
										
						
%
%
%
%


	In order to relate the results of the proposed two-stage QAP based enhancement methods with earlier enhancement methods \cite{journals/pami/HongWJ98,Gottschlich2012,Chikkerur2007198}, EER's are compared in Table \ref{eer}. Table \ref{eer} show the EER for all four datasets of the FVC2004 database with and without using the QAP based enhancement techniques. EER presented for no pre-enhancement, Gabor enhancement and STFT enhancement method are cited from \cite{Fronthale2008} and for ODF enhancement \cite{Gottschlich2012}, EER is obtained using gradient based orientation estimation technique. EERs in Table \ref{eer} show that it is not self-evident that enhancement techniques should improve the recognition performance of the system \cite{Fronthale2008}. On the contrary, it can degrade the recognition performance if the images are noisy. This can be seen in the case of traditional Gabor enhancement and STFT enhancement methods as EER of the verification system is increased for them as compared with EER of original fingerprint images. However, when these techniques are combined with our proposed first stage QAP method, led to lower EERs for all the four datasets. The improvement is specifically pronounced for the traditional Gabor, STFT and ODF enhancement techniques, where significant relative improvement is observed between 1.54\% to 50.62\%. Improved values are marked in bold face. The proposed QAP based enhancement methods improve the EER by as much as 50\% relative to the traditional Gabor/STFT/ODF enhancement methods \cite{journals/pami/HongWJ98,Gottschlich2012,Chikkerur2007198}. No improvement in the EER's of the DB3 and DB4 datasets is observed when using only QAP relative to the original fingerprint images. It shows that using only QAP is not enough to improve the performance for the thermal sweeping sensor dataset (DB3) and SFinGE v3.0 senor dataset (DB4), rather it degrades the system performance for them. However, when QAP is combined with Gabor/STFT/ODF enhancement methods performance is improved for these datasets. The lowest EERs of 9.12\% (QAP + Gabor), 8.27\% (QAP + STFT) and 5.40\% (QAP + STFT) for $DB1-DB3$ datasets are acheived while using QAP based enhancement methods. Lowest EER for fingerprint images of $DB4$ is 7.30\% for no pre-enhanced images. The second lowest EER of 9.32\% is achieved using QAP + Gabor enhancement method for FVC2004 $DB4$.

					\vspace{-2mm}

							\section{Conclusion}\label{Conclusions}
		
		In this paper, a two-stage quality adaptive fingerprint enhancement method is proposed which processes the fingerprint images based on their quality characteristics. The two-stage fingerprint enhancement scheme is preceded by a novel fuzzy c-means clustering based fingerprint quality assessment method to cluster the fingerprint images as dry, normal dry, good, normal wet and wet. Emphasizing the enhancement for fingerprint images of different qualities $Q_c$, a quality adaptive preprocessing method is designed in first stage which controls unsharp mask filter parameter according to quality $Q_c$ of the fingerprint images. In the second stage processing, the enhanced fingerprint images with first stage quality adaptive preprocessing method are further enhanced with Gabor, STFT and ODF enhancement techniques. The experimental results show that the proposed two-stage enhancement utilizing FQA is able to handle various fingerprint quality contexts and achieves better performance in combination with Gabor/STFT/ODF enhancement algorithms. We achieved lowest EERs of 9.12\% (QAP + Gabor), 8.27\% (QAP + STFT) and 5.40\% (QAP + STFT) for FVC2004 $DB1$, $DB2$ and $DB3$ datasets respectively and second best EER of 9.32\% (QAP + Gabor) for FVC2004 $DB4$. These improvement in the results for traditional enhancement techniques can be termed as a paradigm for future studies of quality based fingerprint enhancement.	
			
		A future direction of this study can consider few other frequency and spatial domain features to improve FQA performance. Furthermore, proposed
		FQA based first stage QAP can be combined with other enhancement techniques for better results.


\bibliographystyle{elsarticle-num}

\end{document}